# Impact of Model Interpretability and Outcome Feedback on Trust in AI

Daehwan Ahn
FHCE Department,
University of Georgia, United States
daehwan@uga.edu

Abdullah Almaatouq
MIT Sloan School of Management,
Massachusetts Institute of Technology, United States
amaatouq@mit.edu

Monisha Gulabani
Wharton School,
University of Pennsylvania, United States
monisha.gulabani@gmail.com

Kartik Hosanagar
Wharton School,
University of Pennsylvania, United States
kartikh@wharton.upenn.edu

## ABSTRACT

This paper bridges the gap in Human-Computer Interaction (HCI) research by comparatively assessing the effects of interpretability and outcome feedback on user trust and collaborative performance with AI. Through novel pre-registered experiments (N=1,511 total participants) using an interactive prediction task, we analyzed how interpretability and outcome feedback influence users' task performance and trust in AI. The results counter the widespread belief that interpretability drives trust, showing that interpretability led to no robust improvements in trust and that outcome feedback had a significantly greater and more reliable effect. However, both factors had modest effects on participants' task performance. These findings suggest that (1) interpretability may be less effective at increasing trust than factors like outcome feedback, and (2) augmenting human performance via AI systems may not be a simple matter of increasing trust in AI, as increased trust is not always associated with equally sizable performance improvements. Our exploratory analyses further delve into the mechanisms underlying this trust-performance paradox. These findings present an opportunity for research to focus not only on methods for generating interpretations but also on techniques that ensure interpretations impact trust and performance in practice.

## 1 INTRODUCTION

One of the most important trends in recent years has been the growth of predictive analytics. With advances in machine learning (ML), ML-based artificial intelligence (AI) systems often exceed human-level performance in a variety of domains [69, 85, 92]. Despite the high performance of these systems, users have not readily adopted them [12, 15, 63]. Such reluctance to incorporate algorithms into decision-making has been demonstrated for many years. In a meta-analysis of 136 studies that compared algorithmic and human predictions of health-related phenomena, algorithms outperformed human clinicians in 64 studies (about 47% of the time) and demonstrated roughly equal performance in 64 studies. Human clinicians outperformed algorithms in only eight studies—that is, about 6% of the time [38]. Nevertheless, Grove and Meehl [38] found that algorithms were not widely used in making health-related decisions. Similarly, other studies show that AI has not been widely adopted in medical settings [63], in clinical psychology [93], in firms [83], by professional forecasters across various industries [30], or in a variety of tasks typically performed by humans [12, 15].

To address the issue of algorithm aversion, two primary research streams have emerged within HCI communities: interpretability and performance of ML algorithms. With regard to interpretability, some studies suggest that a lack of interpretability—the ability to explain or present how a model arrives at results in terms understandable to humans—may hinder the adoption of ML algorithms. This is because users may be hesitant to trust a system whose decision-making process they do not understand [7, 46, 73]. Despite this widespread belief, empirical support is relatively limited. Some recent studies have attempted to quantify the impact of interpretability on user trust, but their findings have been inconclusive. The inconsistency in results arises either from small sample sizes, as noted in Panigutti et al. [76], or from conflicting findings. For example, Bansal et al. [5] suggest that interpretability enhances user trust, while Poursabzi-Sangdeh et al. [79] argue the opposite. Adding another layer of complexity, Wang & Yin [96] indicate that interpretability affects trust positively only under certain conditions, varying based on the format of presentations and the expertise level of users. It's also worth noting that while some studies point to interpretability as a factor leading to overreliance on algorithms [18, 52], overreliance should not be conflated with trust, as they are separate albeit related concepts.

In terms of the performance of ML algorithms, existing research suggests that providing information about model accuracy can enhance user trust in algorithmic decision support. For instance, Yin et al. [102] demonstrate that both stated and observed accuracy levels of ML algorithms influence user trust. Similarly, Rechkemmer and Yin [81] compared multiple performance indicators and found that both the stated and observed accuracy had a more significant impact on user trust than the level of model confidence did, although all these indicators positively influenced trust. Additionally, Dietvorst et al. [23] found that users tend to avoid relying on algorithmic decision support after witnessing errors made by these algorithms.

Our paper aims to bridge the gap at the intersection of these two research streams. First, although prior studies have attempted to unveil how either interpretability or performance of ML algorithms influences user trust and human-AI collaborative performance—also referred to as task performance—few have examined these factors in a comparative manner. Thus, there is limited information on which factor has a more substantial impact and whether an interaction exists between these indicators when presented simultaneously. Second, while previous studies have investigated how individual factors like interpretability affect user trust and task performance [5, 23, 76, 79, 81, 96, 102], the relationship between these two outcome measures remains underexplored. Specifically, it remains unclear how increased trust is associated with improved task performance and what mechanisms underlie this relationship. Third, "trust calibration" is frequently employed to measure user trust in algorithmic support [67, 96, 98, 101]. This metric is a composite of user trust and AI performance; for example, user acceptance of an AI suggestion is classified as proper trust if the AI prediction is correct, and as overtrust if the AI prediction is incorrect. While this entangled metric is valuable for assessing complementary performance between humans and algorithms [75, 98], it fails to disentangle the unique impacts of trust from those of model accuracy. This limitation exists because the calibration of user trust—whether manifest as overtrust, undertrust, or proper trust—is not solely determined by the user's intention or behavioral choices, but also depends on the AI's performance.

To fill this gap, our study seeks to understand the influence of interpretability and outcome feedback on users' trust and task performance. Outcome feedback is defined as the post-hoc provision of the actual outcome, intended to confirm the prediction accuracy of both humans and AI for a given event—also known as observed accuracy in the literature [81, 102]. In particular, we assess how interpretability and outcome feedback affect participant trust and performance in a prediction task in order to understand whether these factors increase trust in AI and, if so, which factor has more subtle impact and whether the increased trust is associated with greater human accuracy in the task. We study two levels of interpretability described in the literature, global and local interpretability. Global interpretability clarifies which variables are important to the model's decision-making in aggregate, while local interpretability clarifies which variables are important for a specific decision [73]. Instead of using trust calibration, we employ the weight of advice (WoA) as a measure of behavioral trust to quantify the extent to which users adjust their initial decisions based on AI advice. WoA provides advantages over trust calibration by allowing us to capture varying degree of trust, overtrust, and undertrust, based on users' behavioral choices, while separating out the influence of model accuracy. For additional details on WoA, please refer to the section "Behavioral Trust Measure" and "Choosing The Right Metric: Behavioral Trust vs. Trust Calibration."

In a series of web-based experiments, we investigated how interpretability and outcome feedback affect interactions between humans and AI algorithms. We chose a real-life environment (i.e., interactions with AI advisors) in which lay users can naturally make decisions without any specific training [32]. Our chosen task is also one for which modern AI performs better than humans; if an AI advisor performs worse or equally well compared to humans, there is little benefit to using AI. The main task for our experiments is predicting the outcome of speed dating events with help from a pre-trained ML model [81]. We used a dataset compiled by Fisman et al. [31], which has been used in many related studies [64, 81, 102] to investigate factors affecting trust in ML systems.

Our findings counter the idea that interpretability is a key driver of human trust in AI systems. Specifically, we discovered that neither global nor local interpretability led to robust improvements in trust. In contrast, outcome feedback had a significantly more reliable and greater impact on trust. However, both interpretability and outcome feedback had only minimal effects on task performance. Intriguingly, we observed a paradox: an increase in trust in AI due to outcome feedback did not correspond to proportional improvements in task performance. Through exploratory analyses, we probed the mechanisms underpinning this *trust-performance paradox* associated with outcome feedback. We found that outcome feedback induced users to both overtrust (i.e., overshooting, where users made decisions that exceeded the AI's suggested level of advice) and undertrust (i.e., contradicting, where users chose to go against the AI's advice), thereby compromising human-AI collaborative performance[1]. Our time-dependent analyses further showed that if individuals initially trust an AI system (i.e., adopt its advice in a specific task) but later find that this trust is misplaced (i.e., the AI performs worse than the human's initial prediction in the same task), their trust in the AI significantly diminishes in subsequent tasks. This often leads them to make choices contrary to the AI's advice, further undermining collaborative performance.

Our contributions to the HCI field are outlined below:

- Our study comparatively assessed both the interpretability and performance of ML algorithms, two central themes in HCI related to user trust and collaboration with AI. We found that while interpretability does not substantially enhance trust, outcome feedback significantly and reliably does.
- We scrutinized the relationship between user trust and task performance. To do so, we disentangled user trust from

[1]In this paper, the terms 'overtrust' and 'undertrust' are used differently than in trust calibration contexts. Within the WoA framework, 'overtrust' typically refers to overshooting (i.e., WoA > 1), while 'undertrust' signifies contradicting (i.e., WoA < 0). Conversely, in the trust calibration context, these terms are intertwined with both user adoption of AI advice and the AI's predictive performance on a specific task. Here, 'overtrust' indicates scenarios where users accept incorrect AI advice, and 'undertrust' denotes situations where users do not follow correct AI advice. More details on these two metrics and our rationale for choosing WoA over trust calibration are discussed in Section 6.2.

model accuracy by employing a behavioral trust measure, weight of advice. Our findings revealed a trust-performance paradox influenced by outcome feedback, where increased trust does not result in equivalent gains in task performance.

- Our study shed light on the mechanisms underlying this trust-performance paradox. Specifically, our exploratory analyses discovered that outcome feedback induces users to both overtrust (i.e., overshoot) and undertrust (i.e., contradict) AI, thereby undermining task performance. Additionally, our time-dependent analyses additionally pinpointed when users tend to contradict AI advice, and how this adversely impacts task performance. This result confirms that reliance on AI is not isolated to a single task but is shaped across tasks in a sequential manner.

## 2 RELATED WORK

To address the issue of AI adoption, prior research has delved deeply into understanding factors that affect user behavior and trust in modern AI systems [11, 16, 18, 25, 44, 57, 67, 87, 91, 97, 101, 105]. This stream assessed a broad range of factors, such as human control over algorithmic decisions [24], whose and what type of decision-making is replaced by algorithmic decisions [62, 99], inherent uncertainty in the decision-making domain [22], algorithm transparency [36, 47, 54, 72], and varying levels of information complexity within the model [58].

Within this broader context, two particular areas have received significant focus within HCI research communities: interpretability and performance of ML algorithms. In terms of interpretability, several pioneering studies have attempted to quantify its impact on user trust, but the findings have been inconclusive. For example, Poursabzi-Sangdeh et al. [79] suggested that greater interpretability doesn't necessarily encourage users to rely more on AI predictions compared to black-box models. In contrast, Bansal et al. [5] demonstrated that interpretability enhances the likelihood of users accepting AI advice. Similarly, Panigutti et al. [76] observed that users are more inclined to follow AI recommendations when interpretability features are included in clinical decision support systems. Further complicating the matter, Wang & Yin [96] found that in domains where participants had low expertise, none of the explanation formats improved trust calibration. However, when participants had more domain-specific knowledge, two out of the four explanation formats led to a modest increase in appropriate trust levels.

Regarding the performance of ML algorithms, several studies have highlighted different dimensions that influence user trust. Yin et al. [102] discovered that trust is affected by both the stated and observed accuracy of a model. In a similar vein, Fügener et al. [32] explored how outcome feedback influences users' willingness to delegate tasks to AI systems. Similarly, Dietvorst et al. [23] found that users often refrain from relying on algorithmic decision support if they have witnessed errors committed by the algorithm, underscoring the critical role of performance in shaping user trust. Furthering this understanding, Yu et al. [103] noted that system failures impact trust more significantly than system successes. Expanding on this, Rechkemmer and Yin [81] demonstrated that the model's expressed confidence level does have a significant bearing on user trust, although stated and observed accuracy tend to have a greater impact. Lu & Yin [64] further found that when performance feedback is limited, people often resort to their level of agreement with the model's predictions on specific cases as a heuristic for gauging the model's overall reliability. More recently, He et al. [42] assessed different presentations of stated accuracy (i.e., analogies vs. non-analogies) in relation to trust calibration, finding that analogies alone are not sufficient for achieving appropriate reliance.

Our research diverges from previous studies in three key ways: 1) We distinctively compare both interpretability and performance of ML algorithms to assess their impact on trust in AI and task performance. 2) We disentangle user trust from model accuracy by incorporating a behavioral trust measure, weight of advice, and further investigate the relationship and underlying mechanisms between a user's behavioral trust and task performance. 3) We examine the dynamics of reliance on AI systems, focusing on the sequential interactions between human and AI. From the perspective of product design, it would be useful to understand whether the effects of interpretability differ depending on the presence of feedback about a model's accuracy. Additional clarity on this issue could help connect the two existing streams of research and provide insight into what changes could be made to otherwise capable ML systems in order to improve user trust and adoption.

## 3 EXPERIMENT DESIGN

We ran two web-based experiments, both of which used the same experimental design and prediction tasks and were implemented using the Empirica virtual laboratory platform [2]. However, the user interface differed between the two experiments (see SI Figures S1, S2, S3, S4, and S5) and participants were recruited from different online recruiting panels (Experiment 1: Amazon Mechanical Turk; Experiment 2: Prolific), which allowed us to ensure that the results held across panels and regardless of the specific task presentation. We notably found no significant difference in results between the two different panels and presentation settings.

In our experiments, participants (n = 800 in Experiment 1; n = 711 in Experiment 2) made predictions about the outcomes of speed dating events, first without and then with AI predictions. To assess the impact of model interpretability and outcome feedback on user trust and prediction accuracy, participants were randomized to one of six conditions in a between-subjects experiment design.

### 3.1 Prediction Task

The task, consisting of two phases, asked participants to predict whether couples who had previously met through speed dating would want to pursue a second date.

*3.1.1 Phase One.* The first phase involved 12 task instances (the same instances were used for both experiments). The first two instances were for practice purposes, and participants were informed that the results would not be used in data analysis. These two practice task instances appeared in consistent order for all participants, but the next ten (the results of which were used for data analysis) were randomized. Each task instance presented information about one couple that met through speed dating and asked participants to predict the likelihood that the couple would want a second date. The provided information included (1) demographics (age and race of the man and woman), (2) ratings (the man's and woman's ratings of each other across six attributes: attractiveness, sincerity, intelligence, shared interests, fun, and ambition), and (3) interest correlation (a score representing the similarity between the man's and woman's stated individual interests). Participants made predictions on a slider scale ranging from 0% (extremely unlikely to want a second date) to 100% (extremely likely to want a second date).

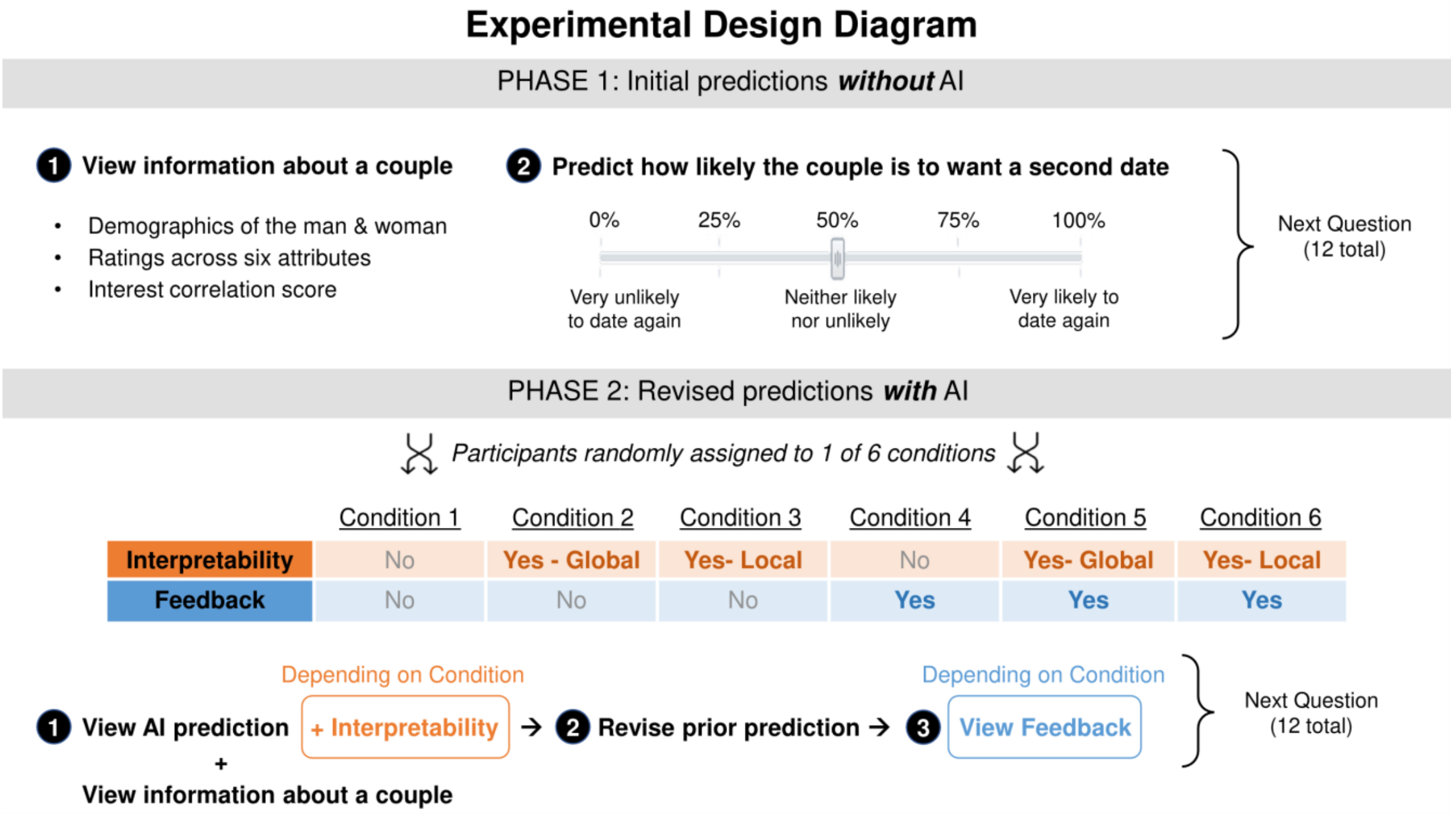

**Figure 1: Experimental Design Diagram**

*3.1.2 Phase Two.* The second phase involved the same 12 task instances. In each of these, participants had an opportunity to revise their prior prediction from phase one after receiving the AI advisor's prediction for that couple. The AI advisor's prediction ranged on a scale from 0% (extremely unlikely to want a second date) to 100% (extremely likely to want a second date). Similar to phase one, participants were informed that the first two task instances were for practice purposes and that only the revised predictions from the remaining ten task instances of phase two would count towards their final score. The task instances in phase two appeared in the same order as they did in phase one.

We chose this prediction task because it is relatable for participants and realistic to how AI is used in the real world (i.e., online dating applications frequently incorporate predictive analytics).

## 3.2 Procedures

All participants received the same information in phase one and the same AI predictions in phase two. However, in phase two, participants received varying levels of interpretability and/or outcome feedback, depending on the condition into which they were randomized. There were three interpretability levels (no interpretability, global interpretability, and local interpretability) combined with two outcome feedback levels (no-feedback and with feedback) for a total of six conditions.

When interpretability was provided, it was delivered alongside the AI prediction so that participants could consider both before making their final prediction in a given task instance. When outcome feedback was provided, it was furnished after participants made their final prediction for a given task instance because the feedback revealed the actual outcome (i.e., whether the couple went on a second date). Nonetheless, because outcome feedback was provided instance by instance, participants could take outcome feedback from prior task instances into account before making future predictions. This format is analogous to common real-world AI interactions with agents such as Amazon Alexa, Google Home, and Apple's Siri. In such interactions, users can observe the accuracy of the agent's understanding of their questions—and oftentimes the accuracy of the agent's response, depending on the kind of question asked (e.g., "What is the weather going to be like today?")—prior to future interactions with the agent.

An illustrative diagram of the experimental design can be found in Figure 1. As described in section 3.1, the experiment consisted of two phases. Phase one involved participants making initial predictions without AI in 12 task instances, with each instance being composed of two steps. In step one, participants viewed information about one couple, and in step two, participants predicted the likelihood that the couple would want a second date. Phase two involved participants revising their initial predictions from phase one after receiving the predictions of an AI system. Phase two also had 12 task instances (each instance corresponded to an instance from phase one), but the steps in each task instance depended on the condition to which a participant was randomized. As described above, there were six conditions that varied in the levels of interpretability and outcome feedback they provided. For all conditions, step one involved viewing the information about the couple, repeated from phase one, and the AI prediction. For conditions that included interpretability, the AI prediction was accompanied by either a global or local interpretation. For all conditions, step two involved revising the initial prediction the user made in phase one. For conditions that included outcome feedback, there was a third step that involved viewing the actual outcome (i.e., whether or not the couple went on a second date).

## 3.3 Description of the Model's Interpretations

The interpretations in this experiment explained what led the AI system to make its predictions, either in aggregate (i.e., global interpretability) or for a specific prediction (i.e., local interpretability) [73]. Global interpretations were extracted using SHAP [65], and local interpretations were extracted using LIME [82][2]. The interpretations were provided as bar charts, a common way of presenting model interpretations. Furthermore, to confirm that participants understood the provided interpretations, they were asked in an exit survey to report the ease with which they understood the information they were given. None of the participants in any of the conditions indicated that they had difficulty understanding the AI system. For additional details regarding the interpretations, see SI Figures S3 and S4. For details regarding the participants' self-reported ease of understanding, see SI section "Self-Report Measures."

## 3.4 Trust and Performance Measures

*3.4.1 Behavioral Trust Measure.* Our measure of behavioral trust is weight of advice (WoA), a measure frequently used in the literature on trust (e.g., trust in AI) and in the literature on advice taking [3, 35, 49, 76, 79, 86]. The WoA measure quantifies the degree to which participants update their response (e.g., predictions made prior to seeing AI predictions) towards provided advice (i.e., the AI prediction). In our experiments, WoA is defined as

$$WoA = \frac{(initial\ prediction - final\ prediction)}{(initial\ prediction - AI\ prediction)}$$

$$If\ \ |AI\ prediction - initial\ prediction| < 0.15,\ WoA = NA$$

The numerator indicates how much the participant's final and initial predictions differ. The denominator takes into account where the participants initially fall relative to the AI prediction. If the WoA equals 1, the final prediction matches the AI prediction; if it equals 0.5, the final prediction is the average of the initial and AI predictions; and if it equals 0, the final and initial predictions are the same. If the WoA is less than 0, the participant moved further away from the AI in their final prediction ("contradicting" the AI); likewise, if the WoA is greater than 1, the participant moved beyond the AI ("overshooting" the AI). A higher WoA indicates greater trust in AI, while a lower WoA indicates less trust.

[2]To ascertain that the results were consistent across different presentations of local interpretability, we selected LIME for its ability to provide range conditions for feature attributions, unlike SHAP, which only offers feature attributions. However, our findings indicated that there were no significant differences between experiments utilizing range conditions (Experiment 1) and those that did not (Experiment 2). Additionally, since there is limited research demonstrating the divergent impacts of SHAP and LIME on user trust in existing literature, we have chosen not to additionally test SHAP for local interpretability.

As noted, we dropped WoA scores when |AI prediction - initial prediction| < 0.15. Because participants could only make selections in increments of 0.05 on the slider scale, it was difficult to make small revisions to match the AI (e.g., if the distance between the initial prediction and AI prediction was 0.1, this revision was difficult to make). Therefore, we interpreted predictions within 0.15 of the AI system as being equivalent to the AI prediction. The 0.15 threshold constitutes a deviation from our pre-registration, so we also tested and confirmed that there were no qualitative changes to the results with thresholds of 0.05 (our pre-registered threshold), 0.1, and 0.2 (see SI section "Robustness Checks").

*3.4.2 Performance Measure.* Our measure of performance is the absolute error of the participant's final prediction, which constitutes a deviation from our pre-registration plan. In the context of our experiments, absolute error is calculated as follows:

$$Absolute\ Error = \ |actual\ value - final\ prediction|$$

$$Wherein\ actual\ value = 1\ if\ the\ couple\ went\ on\ a\ second\ date$$
$$0\ if\ the\ couple\ did\ not.$$

Absolute error can range from 0 to 1. An absolute error of 1 indicates that the participant's final prediction was the exact opposite of the actual dating outcome (0 when the actual outcome was 1 or vice versa). An absolute error of 0 indicates that the participant's final prediction was exactly the same as the actual outcome. Thus, an absolute error closer to 0 indicates greater accuracy while an absolute error closer to 1 indicates less accuracy. We also measured performance using square root error and squared error (see SI section "Robustness Checks").

## 3.5 Hypotheses

We predicted that (1) global interpretability, local interpretability, and outcome feedback would all increase trust in AI, (2) there would be an interaction wherein feedback would be most effective in the absence of interpretability, and (3) global interpretability, local interpretability, and outcome feedback would all increase the accuracy of participants' predictions, owing to the increased trust in AI (per our first hypothesis) and to the fact that AI is on average more accurate in our task than human predictors are. All hypotheses were pre-registered (see SI section "Pre-registered Hypotheses").

## 3.6 Statistical Methods

*3.6.1 Training Process for the AI System.* We used the ensemble tree model XGBoost (eXtreme Gradient Boosting) to determine the AI predictions. This model is known for its superior performance in handling structured data and is popular in the literature [17, 21]. Training the model involved first correcting a class imbalance problem inherent in our dataset. Specifically, our dataset had two classes ("match," meaning the couple went on a second date, and "no match," meaning the couple did not go on a second date). The ratio of "match" to "no match" cases was about 1:4.63 (total observations of 1040 and 4822, respectively). Because there were a significantly higher number of "no match" cases to "match" cases, models would tend to classify the prediction results into the majority class (the "no match" class). Down sampling was used to ensure an equal number of cases in each class (specifically, we randomly sampled 1040 of the 4822 "no match" cases to ensure a 1:1 ratio of "match" to "no match" cases). The model was then trained using 5-folds cross validation. Input data included demographics of each man and each woman, their ratings of the partners they met while speed dating, and each couple's interest correlation score. The task was binary classification, with output data of 1 (match) or 0 (no match). The model's out-of-sample accuracy was about 79%.

*3.6.2 Statistical Analysis.* In our prediction task, each participant was required to complete 12 instances of the task. The initial two instances served as practice, while the remaining 10 instances were utilized for data analysis. We conducted tests for differences across conditions at the task level. To prevent the violation of the i.i.d. assumption, all statistical analyses at the task level were based on linear mixed models that included random effects to account for the nested structure of the data [3, 8]. Linear mixed models are beneficial in situations where data exhibit a clustered pattern, which is evident in our study where individual task responses are nested within each participant (with each participant responding to 10 task cases). All statistical tests were two-tailed.

*3.6.3 Standardized Coefficients.* To enable meaningful comparisons of effect sizes across different condition groups while controlling the effects of various levels of difficulty among task instances, we standardized outcome metrics (e.g., trust, performance) within each task instance. The standardized value of measurement X, measured for task instance i, is defined as

$$X_{i,standardized} = \frac{X_i - \mu_X}{\sigma_X}$$

wherein $\mu_X$ is defined as the mean of X across all instances of the task (for all condition groups) and $\sigma_X$ is the standard deviation. These standardizations not only control the effects of varying levels of difficulty among task instances but also enable meaningful comparisons of effect sizes across tasks of different conditions (e.g., interpretability, outcome feedback).

## 3.7 Participant Recruitment and Compensation

We conducted two experiments, which were conceptual replications of each other, involving different participant groups. The design of both experiments was identical, with participants engaging in the same prediction task (described in the above sections "Experiment Design"). However, the user-interface differed significantly in the two experiments and participants were recruited from different online recruiting panels (Experiment 1: Amazon Mechanical Turk; Experiment 2: Prolific), allowing us to assess whether our results held true with different sets of participants and regardless of the presentation of the task. All participants in both experiments provided explicit consent to participate, and the Institutional Review Board (IRB) and Human Research Protections Program at the university where one of the authors is affiliated approved the consent procedures. Details about participant recruitment for each of the two experiments are described below.

***Experiment 1.*** 800 participants were recruited across 4 days from Amazon Mechanical Turk by posting a HIT for the experiment, entitled "Predict the speed-dating outcomes and get up to $6 (takes less than 20 min)". Participants were required to be at least 18 years of age. To ensure adequate attention on the part of participants, basic attention checks were conducted that were not related to the content of the experiment. Participants that did not pass these attention check questions were not allowed to proceed to the experiment.

***Experiment 2.*** 711 participants were recruited across 4 days on Prolific by posting a study entitled "Predict the speed-dating outcomes and get up to $6 (takes less than 20 min)." Participants were required to be at least 18 years of age. Instead of the basic attention check questions used in Experiment 1, this experiment's attention checks involved substantive questions related to the instructions of the task in order to ensure adequate comprehension of the task itself. These attention check questions were presented in a multiple-choice format, and participants who answered a question incorrectly were told which question was incorrect and were asked to try again until all questions were answered correctly.

In both Experiment 1 and Experiment 2, the payment participants received was dependent on their performance in the task. This approach was designed to encourage active participation, following the methodology outlined by Almaatouq et al. [1]. In Experiment 1, participants received $1 in base pay plus up to $5 of performance-based bonuses. In Experiment 2, participants received $2 in base pay plus up to $5 of performance-based bonuses. The higher base pay in Experiment 2 was due to a base pay requirement of Prolific. The formula used to calculate participant pay was the same for both experiments and is detailed below:

$$base\ payment + 0.5 \times \sum_{I=1}^{N} 1 - (actual\ value - revised\ prediction)^2$$

*Where:*

*base payment = $1 in Experiment 1 and $2 in Experiment 2*
*N = number of prediction rounds*
*actual value = 1 if the couple went on a second date &*
*0 if the couple didn't go on a second date*

MTurk Worker IDs and Prolific IDs were automatically collected, and participant data was linked to the IDs for the purposes of participant compensation. Because our study required an interactive experiment system and an incentive compatible system, and because it is currently not possible to create an incentive compatible interactive experiment entirely through MTurk or Prolific, we created our own experiment system that works with MTurk and Prolific. As such, our system needed to collect MTurk Worker IDs and Prolific IDs and link these IDs with participant data in order to calculate compensation for each participant, as compensation was tied to performance in the task. IDs were only used for payment purposes, were deleted after payments were successfully delivered, and were not used in data analysis. The need to collect Mturk Worker IDs/Prolific IDs and link them to participant data was disclosed to and approved by the Institutional Review Board at the university where one of the authors is affiliated.

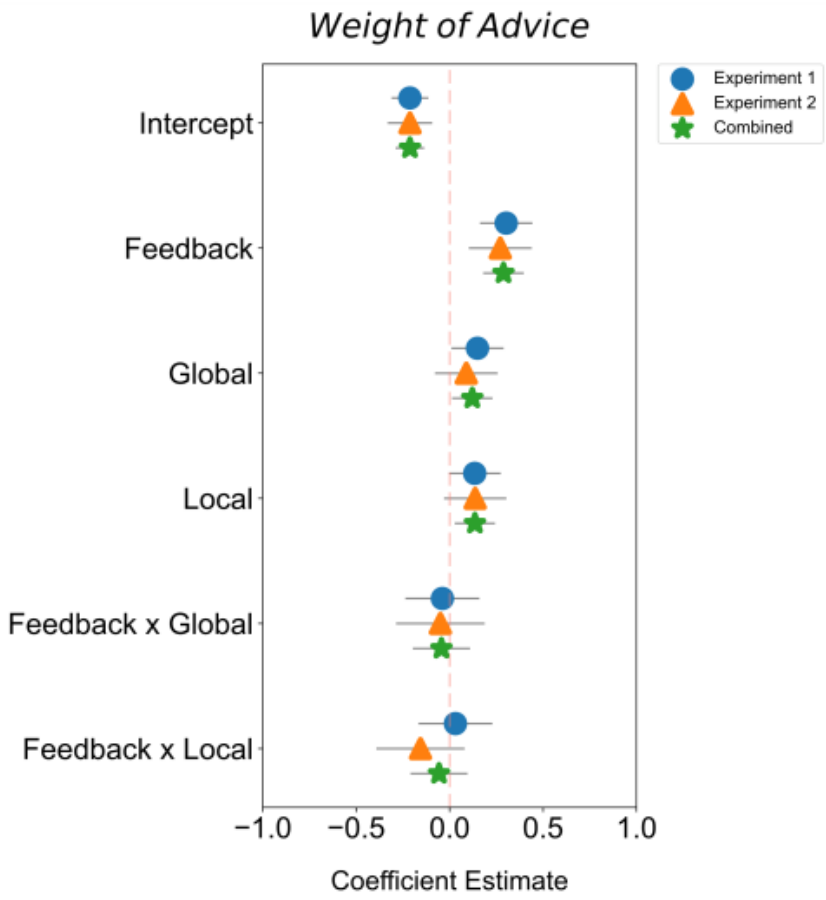

**Figure 2: The Effect of Outcome Feedback and Interpretability on Behavioral Trust**

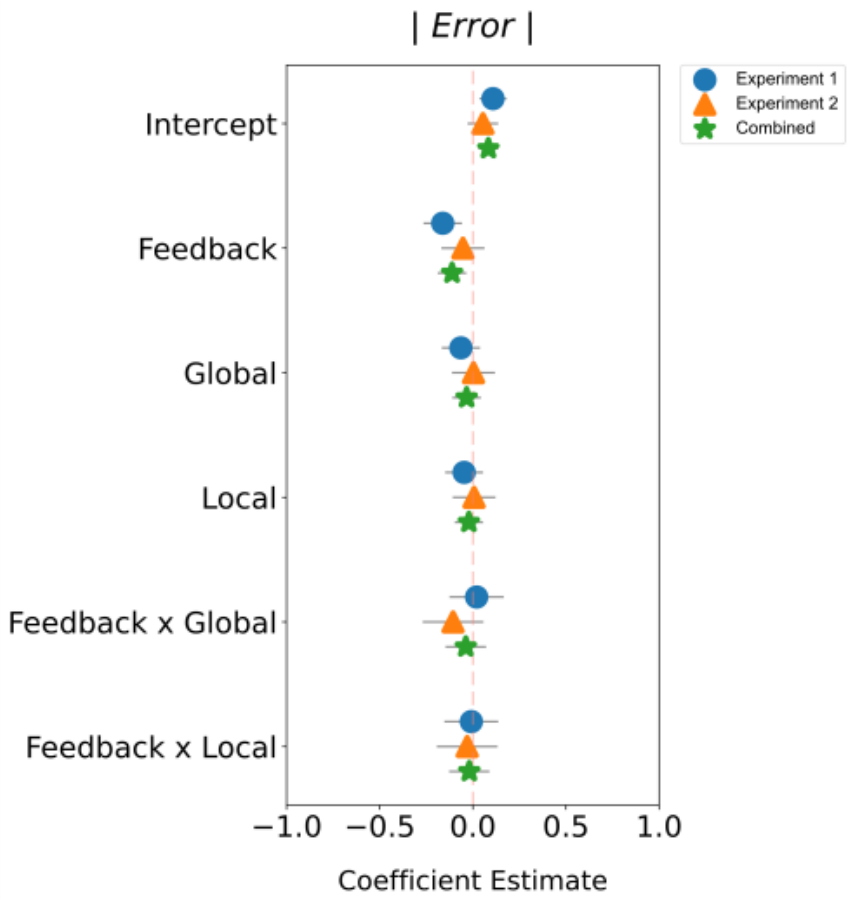

**Figure 3: The Effect of Outcome Feedback and Interpretability on Performance**

# 4 RESULTS

## 4.1 Study 1: Impact of Feedback and Interpretability on Behavioral Trust in AI

This experiment sought to assess the impact of outcome feedback and interpretability on behavioral trust in AI. Figure 2 compares the standardized (i.e., z-scored within each task instance) effect of outcome feedback, interpretability, and the interaction of these two factors on behavioral trust. Behavioral trust was assessed using the WoA metric, as described in the "Trust and Performance Measures" section.

As shown in Figure 2, outcome feedback led to the greatest and most reliable increase in behavioral trust (Experiment 1: $P < 0.001$; 95% CI = [0.163, 0.443]; Experiment 2: $P < 0.003$; 95% CI = [0.103, 0.439]). However, global and local interpretability were not observed to have a robust effect on trust (Experiment 1: global: $P < 0.038$; 95% CI = [0.009, 0.289]; local: $P < 0.059$; 95% CI = [−0.004, 0.273]; Experiment 2: global: $P < 0.298$; 95% CI = [−0.078, 0.257]; local: $P < 0.109$; 95% CI = [−0.030, 0.304]). Furthermore, the effect of interpretability on trust was modest relative to the effect of outcome feedback and not robust to different choices of |AI prediction – initial prediction| thresholds in our definition of WoA (see SI section "Robustness Checks"). Additionally, there was no difference between global and local interpretability in terms of impact on trust (Experiment 1: $P < 0.838$; 95% CI = [−0.154, 0.125]; Experiment 2: $P < 0.575$ ; 95% CI = [−0.118, 0.214]). Contrary to our hypothesis, there was no observed interaction between outcome feedback and interpretability, meaning that feedback and interpretability did not appear to complement (or substitute) each other when provided together (Experiment 1: feedback × global: $P < 0.703$; 95% CI = [−0.237, 0.159]; feedback × local: $P < 0.757$; 95% CI = [−0.166, 0.229]; Experiment 2: feedback × global: $P < 0.684$; 95% CI = [−0.286, 0.188]; feedback × local: $P < 0.197$; 95% CI = [−0.392, 0.080]).

The finding that outcome feedback resulted in the greatest and most reliable increase in trust is not only counter to our hypothesis but also counter to the current focus on interpretability as a central driver of trust in AI systems.

## 4.2 Impact of Feedback and Interpretability on Performance Accuracy

This experiment also sought to assess the impact of outcome feedback and interpretability on participants' performance in the prediction task. Performance was assessed using the absolute error metric, as described in the "Trust and Performance Measures" section. Decreased absolute error indicates improved performance accuracy while increased absolute error reflects the opposite.

Figure 3 compares the standardized effect of outcome feedback, interpretability, and the interaction of these two factors on performance to assess what impact, if any, these factors had on participant performance, beyond the effect that was attributable to the AI predictions themselves.

As shown in Figure 3, outcome feedback led to a further improvement in performance (i.e., decrease in absolute error) beyond that which was attributable to the AI predictions, although this effect was slightly smaller and not significant in Experiment 2 (Experiment 1: $P < 0.003$; 95% CI = [−0.265, −0.058]; Experiment 2: $P < 0.369$; 95% CI = [−0.169, 0.063]). The performance improvement resulting from outcome feedback was consistent with our predictions.

However, contrary to our expectations, neither global nor local interpretability were found to impact performance (Experiment 1: global: $P < 0.224$; 95% CI = [−0.167, 0.039]; local: $P < 0.369$; 95% CI = [−0.148, 0.055]; Experiment 2: global: $P < 0.954$; 95% CI = [−0.112, 0.119]; local: $P < 0.910$; 95% CI = [−0.108, 0.122]). Furthermore, the interaction between feedback and interpretability was not found to impact performance (Experiment 1: feedback × global: $P < 0.784$; 95% CI = [−0.125, 0.166]; feedback × local: $P < 0.913$; 95% CI = [−0.153, 0.137]; Experiment 2: feedback × global: $P < 0.203$; 95% CI = [−0.269, 0.057]; feedback × local: $P < 0.716$; 95% CI = [−0.193, 0.133]).

The finding that outcome feedback improved participant performance in the prediction task, while interpretability was not observed to improve performance, is in line with the previously discussed finding that outcome feedback had a more significant effect on trust in AI than interpretability had. However, it is critical to note that while outcome feedback led to improved performance, the size of that performance increase was relatively small compared to feedback's increase in behavioral trust. Similarly, interpretability was not observed to have an impact on performance in the prediction task, though it was found to increase trust in AI to some extent. This suggests that the relationship between trust in AI and performance in the prediction task may not be as direct as initially assumed. In particular, these findings challenge the assumption that increased trust in AI directly leads to improvements in performance. Instead, this experiment found that improved trust in AI is not always associated with equally sizable performance improvements.

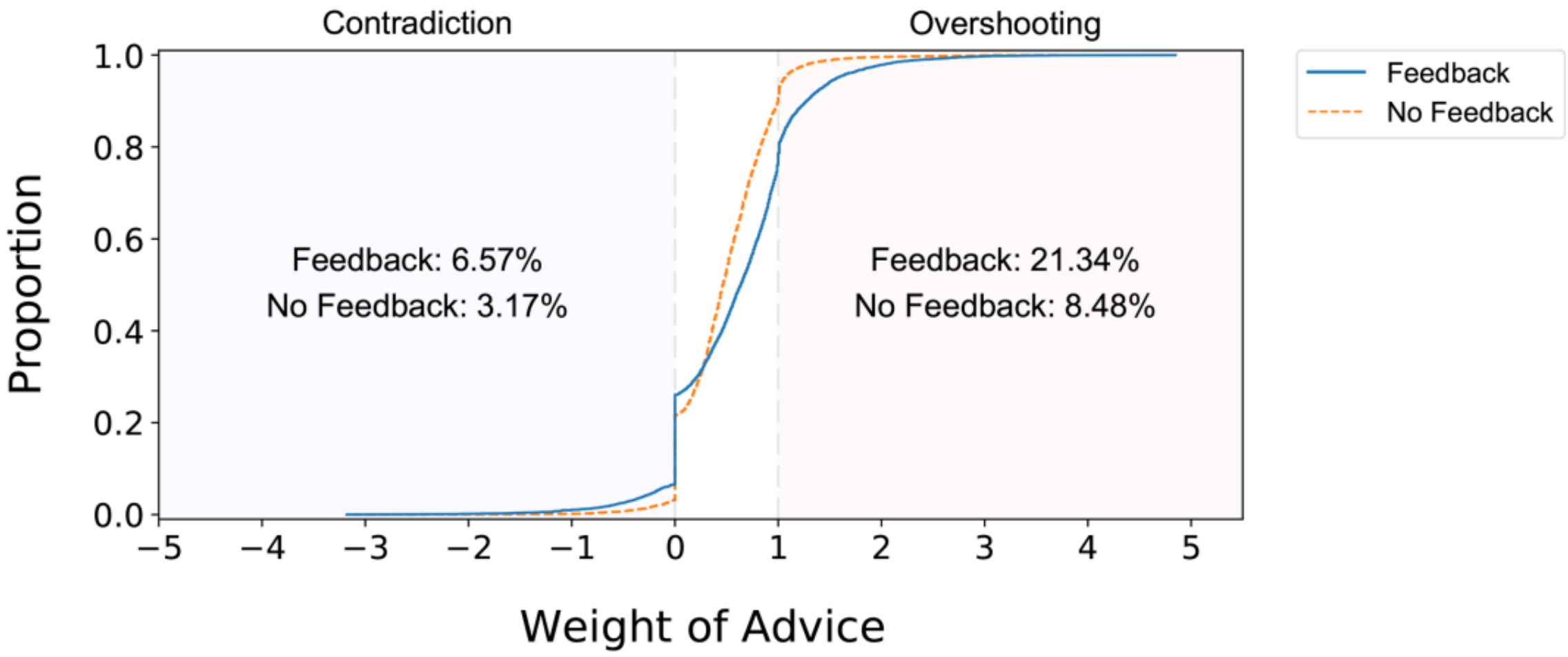

**Figure 4: The Effect of Outcome Feedback on Behavioral Trust Patterns**

## 5 EXPLORATORY ANALYSES

Through exploratory analyses, we sought to answer why the increased trust from outcome feedback is not associated with equally sizable improvements in performance. In particular, we address this paradox from the perspective of users' overtrust and undertrust in AI—factors that have been shown to undermine human-AI collaborative performance [4, 10, 48, 75, 89]. Building upon this, we further investigated whether, when, and how outcome feedback makes users overtrust and undertrust AI and if it further harms human-AI collaborative performance.

In particular, we first show that outcome feedback induces users not only to trust AI more but also to overtrust and undertrust AI more. Then we show that increased overtrust and undertrust undermine human-AI collaborative performance. Additionally, we demonstrate that users contradict AI after their trust in AI backfires, which significantly harms performance in regard to users' time-dependent behavioral trends.

### 5.1 Why is the Increased Trust from Outcome Feedback Not Associated with Equally Sizable Improvements in Performance?

*5.1.1 Outcome feedback simultaneously induces users to overtrust and undertrust AI.* Our next analysis sought to assess why the increased trust from outcome feedback is not associated with equally sizable improvements in performance. Figure 4 compares the empirical cumulative distribution function (ECDF) regarding trust in AI according to the presence or absence of feedback. The x-axis refers to participants' behavioral trust patterns (i.e., WoA) at a task instance-level, and the y-axis represents the cumulative proportion of observations.

As shown in Figure 4, outcome feedback simultaneously induced users to overtrust (i.e., overshooting where WoA > 1) and undertrust (i.e., contradicting where WoA < 0) the AI system's advice. In particular, outcome feedback resulted in a near tripling of overshooting (i.e., from 8.48% to 21.34%) and a near doubling of contradiction (i.e., from 3.17% to 6.57%), relative to the condition when outcome feedback was not given. Also, the feedback group has fewer observations in the range where WoA is between 0 and 1. Taken together, our results show that outcome feedback induced users to make extreme behavioral trust choices (i.e., more extreme WoAs toward both positive and negative directions), resulting in higher variance in WoA distribution.

We statistically tested our findings through the two-sample Anderson-Darling test, which is widely used to compare cumulative distributions while detecting differences at the tail ends of distributions more reliably; in our case, contradiction and overshooting correspond to the tail ends [27]. We confirmed that the feedback and no-feedback groups have different proportions in distributions (i.e., the distribution of the feedback group has higher variance), and this is statistically significant ($P < 0.001$).

*5.1.2 Overtrust and undertrust undermine human-AI collaborative performance.* This analysis sought to assess whether overtrusting and undertrusting hurt human-AI collaborative performance. Reduction in error was used as a measure of human-AI collaborative performance: a positive value means performance improved (i.e., error decreased) after being exposed to AI advice, when compared to a participant's initial prediction. Figure 5 shows how the benefit of AI advice changes according to the different degrees of WoA. On the x-axis, we group task instances according to their WoA values in increments of 1. The y-axis indicates the average reduction in error per group. As shown in Figure 5, reduction in error (i.e., performance improvement) is concave in WoA: increased WoA produced improvements in decision performance initially (i.e., 0 <= WoA < 2), but beyond a point (i.e., WoA >= 2), the benefits dropped and a further increase in WoA only had a negative effect on performance. Similarly, as WoA goes below zero (i.e., participants contradicted AI advice), the further decrease in WoA only hurt human-AI collaborative performance. Notably, as WoA moves toward either extreme in the positive or negative direction, it harms the benefit of AI advice by a larger margin.

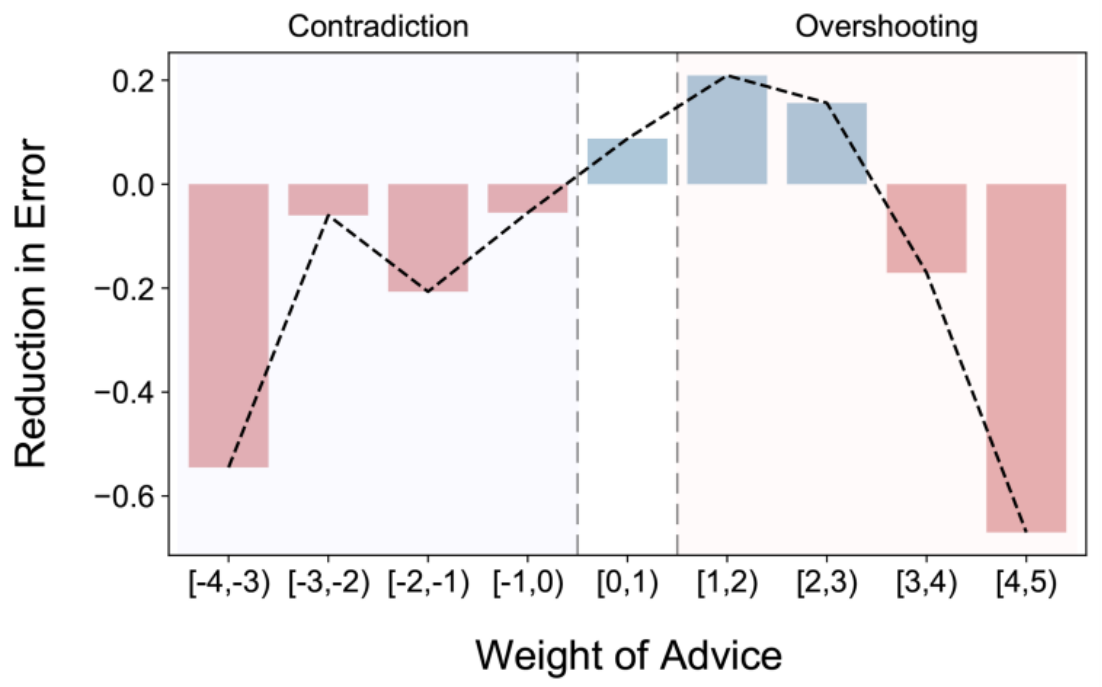

**Figure 5: The Effect of WoA on Human-AI Collaborative Performance**

**Table 1: A Concave Relation between WoA and Reduction in Error**

| Independent Variables | Coef. | 95% CI |
|---|---|---|
| Feedback | 0.081*** | [0.042, 0.120] |
| WoA | 0.799 *** | [0.749, 0.848] |
| WoA2 | -0.193 *** | [-0.220, -0.165] |
| Constant | -0.358 *** | [-0.390, -0.326] |

Note: * p<0.05; ** p<0.01; *** p<0.001.

This result is also supported by our statistical test. Table S1 shows the result of a regression model that tests the relationship between WoA and reduction in error. As shown in Table S1, WoA has a statistically significant quadratic relationship with reduction in error and the coefficient is negative—a concave relation.

This finding was also directly associated with overall performance. Figure 6 compares the ECDF of the final performance according to different feedback conditions. The x-axis represents the absolute error, and a lower value means participants had better performance after human-AI collaboration. The y-axis refers to the cumulative proportion of observations.

As shown in Figure 6, the feedback group has more cases with small errors compared to the no-feedback group. This is what we expect given the increase in WoA. However, despite this benefit, the feedback group also has a greater number of "failures" (i.e., cases having large errors). This double-edge effect may help explain why the increased trust from outcome feedback is not associated with equally sizable improvements in performance. When feedback is given, the increased number of failures (i.e., large errors) offsets the benefit from the increased number of successes (i.e., reduction in error).

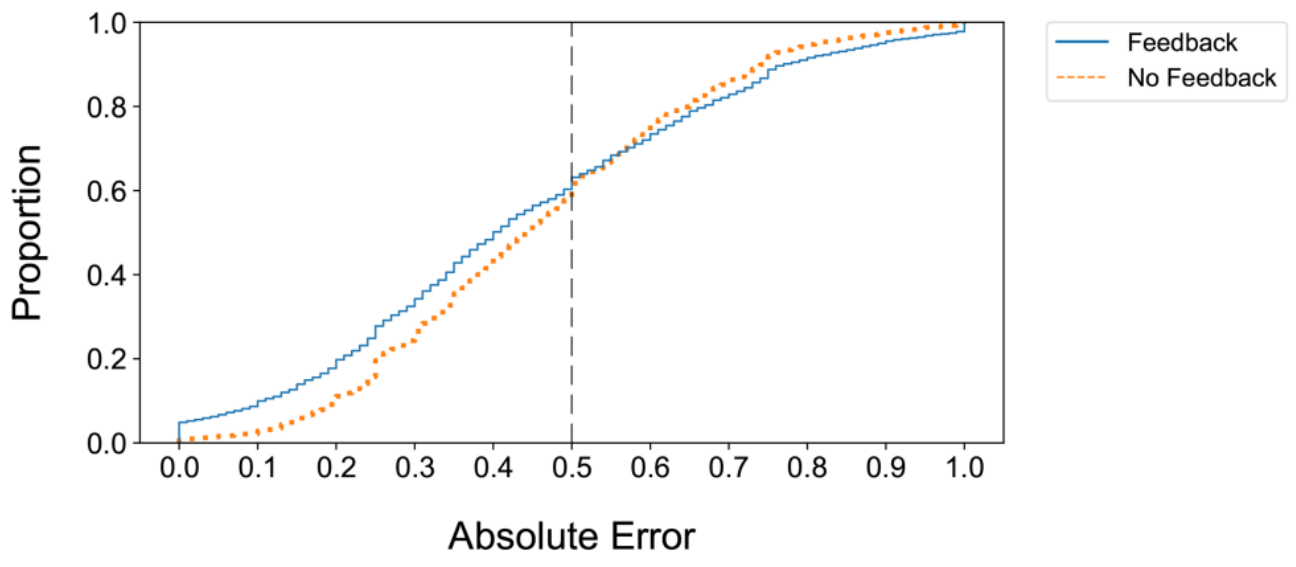

**Figure 6: The Effect of Outcome Feedback on Human-AI Collaborative Performance**

Taken together, these exploratory analyses suggest that outcome feedback induces users to both overtrust AI decision support and to undertrust it. While overtrusting is consistent with a higher trust (increase in WoA), it does not necessarily drive improved performance (reduction in errors). This is in line with the results of prior works that explain the noisy nature of the relationship between trust in AI and performance by grouping "trust calibration" into overtrust (i.e., following the AI system's advice when it is incorrect), appropriate trust (i.e., following its advice when it is correct), and undertrust (i.e., not following its advice when it is correct) [75, 89]. However, we differ from prior research because we disentangle trust from performance, whereas trust calibration is a composite measure of the two. We further explore the mechanism of overtrust and undertrust in the following sections.

## 5.2 When Does Outcome Feedback Induce Users to Overtrust and Undertrust AI?

Our next analysis sought to assess when outcome feedback induces users to undertrust AI decision support more (compared to when feedback is not given), particularly in regard to users' time-dependent behavioral trends. Participants had sequential interactions with an AI advisor, meaning that they received AI-based predictions in addition to interpretability and/or outcome feedback following each task instance. This raises the question of whether a time-dependent trend exists in terms of how these factors affect trust in AI and performance in the prediction task. Exploratory analysis suggests that outcome feedback appears to impact trust and performance over time. Specifically, trust and performance appear to depend on the kinds of experiences a participant had with the AI system in prior task instances.

To start off, we investigate participants' time-dependent trends specific to behavioral trust. Participants' initial predictions were, on average, less accurate than the AI predictions, meaning that participants would have improved their performance if they had trusted the AI. However, there were also instances where a participant's initial prediction was more accurate than the AI prediction; if participants had trusted AI in those cases, their performance would have declined. Because outcome feedback was provided after each task instance, participants knew whether following AI in prior task instances helped or hurt their performance before proceeding to subsequent task instances.

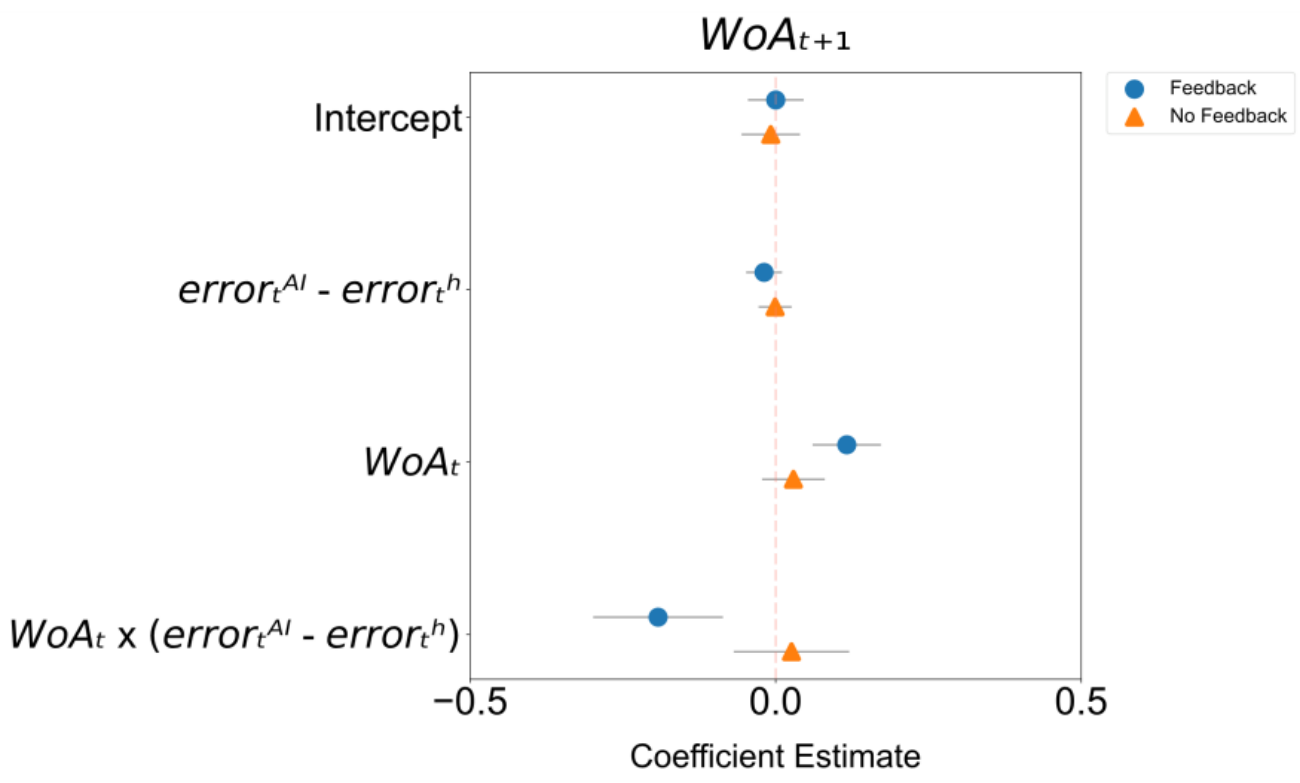

**Figure 7: Time-Dependent Trends Specific to Behavioral Trust**

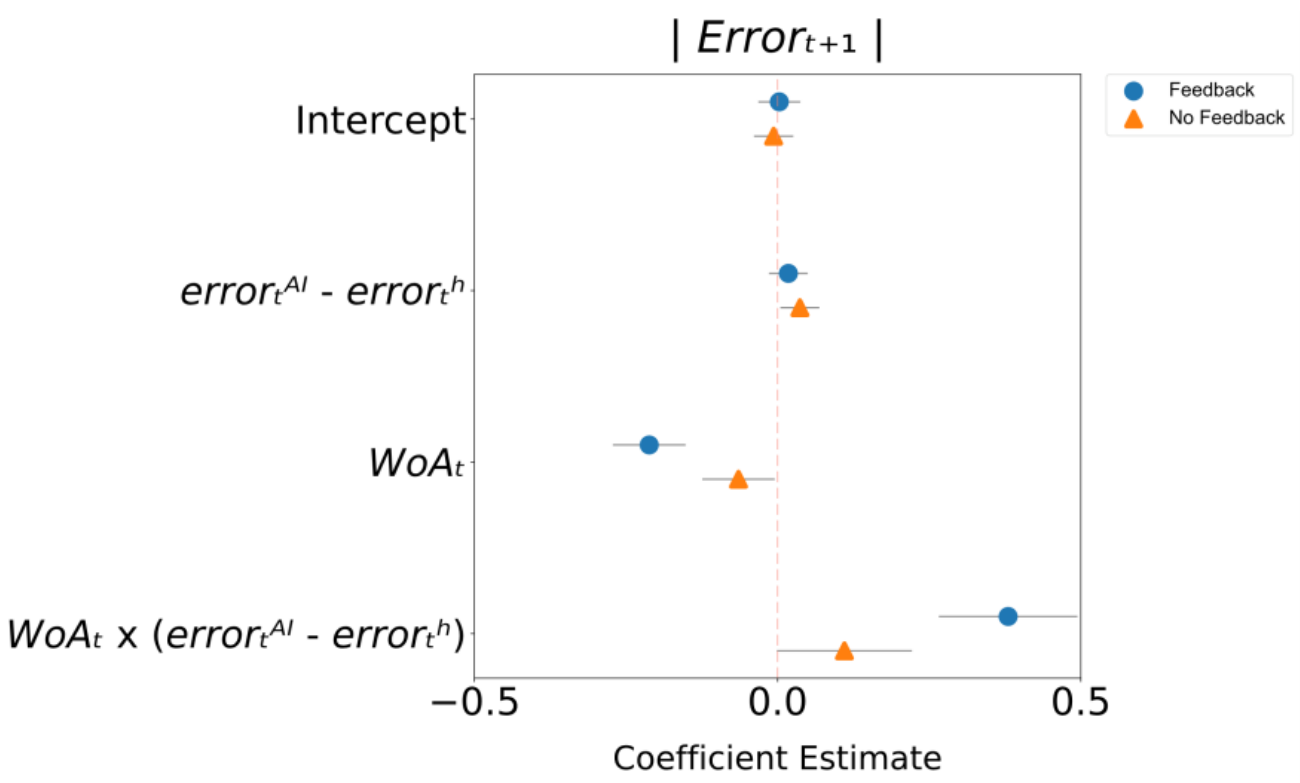

**Figure 8: Time-Dependent Trends Specific to Performance**

Figure 7 compares the standardized effect of three aspects of a given task instance (at time t) on behavioral trust in a subsequent task instance (at time t+1). The first factor ($error_t^{AI} - error_t^h$) represents the initial difference in performance between an AI system and a user in a given task instance. A positive value indicates that the human's initial prediction outperformed that of the AI. The second factor ($WoA_t$) represents the user's behavioral trust in a given task instance. The third factor [$WoA_t \times (error_t^{AI} - error_t^h)$] is the interaction of the first two factors. One scenario this interaction captures is where a participant's initial prediction is more accurate than the AI prediction, but the participant revises their prediction towards the AI prediction, thereby reducing their accuracy (e.g., the AI system's advice "harmed" the participant's performance). For each factor, $WoA_{t+1}$ was compared for the feedback group (all cases in which participants received outcome feedback, combined across both experiments) and the no-feedback group (all cases in which participants did not receive outcome feedback, combined across both experiments).

As shown in Figure 7, WoA was greater at time t+1 as compared to time t for the feedback group (Feedback Group: $P < 0.001$; 95% CI = [0.061, 0.172]; No-Feedback Group: $P < 0.268$; 95% CI = [−0.022, 0.080]). This suggests that, overall, feedback increases trust over time, as seeing feedback for one task instance (time t) tends to increase behavioral trust in the AI advisor in the subsequent instance (time t+1). However, a markedly different, though still time-dependent, effect was observed in cases where following the AI advice "harmed" the participant [reflected in the interaction term $WoA_t \times (error_t^{AI} - error_t^h)$]. In these cases, $WoA_{t+1}$ was significantly reduced relative to $WoA_t$ for the outcome feedback group (Feedback Group: $P < 0.001$; 95% CI = [−0.298, −0.086]; No-Feedback Group: $P < 0.592$; 95% CI = [−0.069, 0.120]). This suggests that the experience of trusting an AI advisor and having one's performance decrease as a result leads to a loss of trust in that system's advice in the subsequent instance. This proposed trend is in accordance with prior research that users do not trust algorithms after observing them fail [23].

These observations that outcome feedback tends to increase trust over time in aggregate but decrease trust after a particular negative experience are consistent with the theory that outcome feedback impacts behavioral trust over time. These trends were only observed for the feedback group, which was expected given that the no-feedback group did not receive information about actual outcomes and thus could not know whether following the AI advice was helping or hurting their performance over time.

Next, we explore participants' time-dependent trends specific to performance accuracy. Similar time-dependent trends were observed regarding the impact of outcome feedback on performance. The factors assessed in Figure 7 are again evaluated in Figure 8, with Figure 8 comparing the standardized effect of these factors on performance (i.e., absolute error) at time t+1.

As shown in Figure 8, there are two observations regarding absolute error at time t+1 that can be analyzed in conjunction with those displayed in Figure 7.

Figure 8 suggests that $|Error_{t+1}|$ was reduced for $WoA_t$ for both the outcome feedback and no-feedback groups, although the effect is very small and not significant for the no-feedback group (Feedback Group: $P < 0.001$; 95% CI = [−0.272, −0.152]; No-Feedback Group: $P < 0.036$; 95% CI = [−0.124, −0.005]). Thus, it appears that when participants trusted AI in one task instance, they tended to have smaller errors (i.e., improved performance) in the next instance, an effect that was stronger when feedback was provided. When analyzed in conjunction with Figure 7, this suggests that when participants in the feedback group trusted AI in one task instance, trust increased even further in the next instance (Figure 7), and this increase in trust was associated with a performance improvement (Figure 8).

Figure 8 also suggests that for the outcome feedback group, $|Error_{t+1}|$ increased for the interaction term $WoA_t \times (error_t^{AI} - error_t^h)$ (Feedback Group: $P < 0.001$; 95% CI = [0.266, 0.494]; No-Feedback Group: $P < 0.053$; 95% CI = [−0.001, 0.222]). Thus, it appears that if trusting AI "harms" a user in one task instance, their error increases in the next task instance. Taken together, Figures 7 and 8 suggest that after an AI system "harms" a user, trust decreases in the next instance (Figure 7), and this loss of trust is associated with reduced future performance (Figure 8). These observations are robust to other operationalizations and performance measures (see SI section "Robustness Checks").

These exploratory analyses suggest that outcome feedback has two time-dependent effects: it generally increases trust and performance over time but can sometimes reduce trust and performance. Specifically, trust in an AI system increases over time when users observe that system performs accurately over time. Nevertheless, AI can at times be more erroneous than the human decision-maker even though it outperforms humans on average. We observe that when humans trust AI but that trust backfires (i.e., AI performs worse than the human in a particular instance), then trust in that AI system drops in subsequent task instances. This drop in trust hurts the human's future performance and limits users from fully extracting the potential value of AI decision support. Research that specifically studies these time-dependent effects and research that seeks to understand the relationship between trust in AI and performance in prediction tasks will be important extensions of the literature.

# 6 DISCUSSION AND DIRECTIONS OF FUTURE RESEARCH

## 6.1 A Step toward Understanding and Fostering Appropriate Trust in AI Systems

Fostering appropriate trust in AI systems is challenging, primarily due to their inherent complexity. To gain a deeper understanding of reliance on AI systems, it is crucial to consider multiple facets simultaneously. These encompass algorithm-related factors such as model accuracy [42, 64, 102], explainability [5, 18, 76, 79], controllability [24], uncertainty [22], and information complexity [58]. Equally important are user-side aspects, which include human cognition, subjective and psychological perspectives of users [18, 20, 39, 59, 94, 96, 104, 106], as well as diverse levels of expertise and literacy in both AI and the relevant tasks [26, 55, 90]. Additionally, the unique characteristics of interactions between the algorithm and users should also be considered. This includes factors such as the consistency between algorithm and user decisions [70], and the sequential interactions between them [74, 88]. Lastly, socio-contextual factors and task characteristics surrounding the AI system, including economic motivations [28, 29], organizational contexts [78], and consumer perspectives [99], could significantly contribute to this dynamic.

Considering the inherent complexity of the subject, our study aligns with the ongoing efforts to expand research dimensions. We have shifted our focus from the time-invariant impact of single factors to a more dynamic examination of the time-dependent and comparative impacts between multiple factors, which include different types of interpretability and outcome feedback.

Our time-dependent analysis revealed that reliance on AI is not isolated to a single task but is shaped across tasks in a sequential manner. This aligns with recent studies, highlighting significant insights that human-AI interaction evolves over time. For example, recent research has demonstrated that the development of trust in AI systems progresses over multiple sessions, with the initial impression of AI performance playing a crucial role in shaping users' perceptions of these systems [74, 88]. While numerous experimental studies on AI reliance have been conducted in settings with sequential tasks, they often focus on the aggregated level impacts of factors in human-AI interaction. In contrast, our study identifies specific instances where humans contradict AI advice and examines the effects on human-AI collaborative performance. By extending our analysis to include time-dependent interactions between humans and AI, we provide valuable insights that can significantly enhance the development of appropriate trust in human-AI collaboration.

The literature identifies two broad categories of factors influencing trust in AI: performance-based (such as overall accuracy and outcome feedback) and model-based (such as interpretability and transparency). Our paper specifically focuses on a comparative examination of interpretability and outcome feedback, assessing their impacts on trust in AI and performance. The rationale for this comparative study design is twofold. Firstly, from a practical implications perspective, such studies can provide guidelines for the design and selection of factors in scenarios where both elements coexist, especially in the context of limited resources. Practitioners can selectively choose factors for AI system design, depending on the objectives of the service, while considering the relative impact size and ease of incorporation. Secondly, a comparative study allows for an examination of interactions among different factors. Although our study did not observe interactions between the presence of interpretability and outcome feedback, recent research has revealed meaningful interactions among factors in AI reliance. For example, Kahr et al. [51] found that a specific type of explanation (i.e., human-like explanations) did not independently affect user trust in AI systems, but it did have an interaction effect with model accuracy—human-like explanations boosted trust in high-accuracy models. Future research could conduct extensive comparisons ('horse races') between various factors hypothesized to influence trust, along with their interactions, across many different tasks or situations.

## 6.2 Choosing The Right Metric: Behavioral Trust vs. Trust Calibration

The choice of metric acts as a lens through which phenomena are examined, shaping the structure and details of the study. Selecting the most appropriate metric, with careful consideration of the research question and objective, is a critical step in study design. In research on trust and reliance in AI systems, two primary types of metrics are commonly used: behavioral trust (e.g., WoA) and trust calibration (e.g., appropriate reliance). These metrics are grounded in different philosophies and possess their own unique advantages and disadvantages. Our study chose the behavioral trust metric (specifically WoA) over trust calibration as the primary metric. This decision was based on the following rationale.

Firstly, as outlined in the introduction, one of our key objectives is to explore the relationship between user trust and task performance in human-AI collaboration. Trust calibration, while insightful, is a complex metric that intertwines user trust with the task performance of the model. To address this complexity, we employed WoA, a behavioral trust measure, to disentangle user trust from model accuracy. This strategic choice enabled us to identify a trust-performance paradox influenced by outcome feedback within a two-stage setting. The first stage involved examining the impact of interpretability and outcome feedback on behavioral trust. The second stage focused on the relationship between improved trust and task performance, as well as the underlying rationale behind this correlation.

Secondly, WoA provides the concepts of overshooting and contradiction, which are pivotal in elucidating our key findings and the mechanisms underlying them. These concepts specifically address users' unpredictable behaviors in response to AI support, setting them apart from the notions of overtrust and undertrust found in trust calibration. Overtrust and undertrust in the context of trust calibration are entangled metrics, intricately combining user decision-making and model performance. In contrast, overshooting and contradiction as defined within WoA, provide a more nuanced understanding of how users interact with AI. They go beyond the simple binary of trust and distrust seen in trust calibration, capturing complexities, such as the degree of AI advice adoption, and sometimes paradoxical nature, exemplified by overshooting and contradiction, of user responses to AI recommendations. This distinction is crucial as it allows for a deeper exploration of user behavior patterns that are not readily apparent in the trust calibration model. By leveraging these unique concepts, our adoption of WoA provides a more comprehensive and detailed perspective for viewing and interpreting the dynamics of human-AI interaction.

Thirdly, trust calibration operates on the assumption of complementary performance in human-AI collaboration. For appropriate reliance, it is essential that humans are able to discern when to trust and when to distrust AI. This means selectively adhering to AI decisions when they are likely to be correct, and disregarding them when they are likely to be erroneous [18]. However, a major challenge in predictive analytics is the difficulty in determining when AI predictions are right or wrong. For instance, even an AI model with 90% accuracy fails in 10% of cases, but predicting which cases will fall into this 10% is challenging. Despite efforts to address this, such as through uncertainty modeling, the issue of uncertainty in predictive analytics remains somewhat inherent [33, 34, 68]. In this context, numerous empirical studies have not observed this complementary performance in real-world scenarios due to the difficulty humans face in accurately determining when AI is right or wrong [6, 13, 18, 37, 56, 61, 66, 79, 95, 100, 107]. Furthermore, AI is being increasingly deployed in complex tasks that surpass human cognitive capabilities, suggesting scenarios where AI may outperform both human-only and human-AI collaborative efforts. Given these reasons, the notion of having users completely trust and follow AI decisions is gaining importance. From this practical standpoint, since WoA does not assume the necessity of human-AI complementary performance, it offers additional insights beyond what trust calibration can provide. WoA is particularly relevant in scenarios where complete reliance on AI might be imperative, especially in situations where AI's capabilities significantly surpass those of humans.

Lastly, the use of a disentangled metric facilitates the development of more effective strategies due to its simplicity and better controllability. In the case of trust calibration, the aim to foster appropriate reliance in AI is dual-faceted: it involves not only persuading users to trust and follow AI decisions, but also ensuring that the AI provides accurate predictions. However, controlling AI performance on an individual case basis (i.e., discerning when AI is right or wrong for each case) is challenging. In contrast, influencing user trust is more feasible. Thus, WoA provides an avenue to initiate discussions from the more manageable user perspective, and then to incrementally broaden the scope to more comprehensive solutions. For instance, our study highlighted that users' overshooting and contradiction in AI negatively impacts their collaborative performance, elucidating the specifics of when and how this overshooting and contradiction occurs. These insights are invaluable for future research aimed at exploring methods to mitigate overshooting and contradiction, ultimately enhancing human-AI collaboration. This nuanced approach is not achievable with trust calibration, including more granular versions like Relative Positive AI Reliance (RAIR) and Relative Positive Self-Reliance (RSR), as these metrics still remain closely tied to task performance of model [84]. In this light, the concept of appropriate reliance in trust calibration is viewed as a consequentialist goal, focusing more on the ideal end-state of human-AI collaboration rather than on the gradual process of problem-solving itself.

We selected WoA over trust calibration not because WoA is inherently superior, but because it more closely aligns with the specific research questions and objectives of our study. Each metric has its distinct advantages: trust calibration excels at granularly defining the ideal state of human-AI collaboration (e.g., appropriate reliance, RAIR, RSR), while WoA offers a deeper exploration of human behaviors, encompassing even those that are unreasonable or paradoxical. We strongly advocate for future studies to integrate both behavioral trust and trust calibration metrics, as this combined approach has the potential to yield synergistic insights and foster a more comprehensive understanding of human-AI interactions. Additionally, trust calibration can be further highlighted through research that focuses on the mechanisms of 1) how and why certain factors enhance a user's task-related knowledge, and 2) how this improved knowledge leads to more effective filtering of AI advice for appropriate reliance, consequently aiding in achieving complementary performance between humans and AI systems. For instance, Chen et al. [18] implemented a think-aloud, mixed-methods study to investigate the human intuitions in the decision-making process when adopting AI advice. Their findings provided valuable insights, clarifying why feature-based explanations lead to overreliance on AI, while example-based explanations are particularly effective in fostering complementary human-AI performance.

## 6.3 Why Does Outcome Feedback Affect Trust More Than Explanations Do?

We found that interpretability does not significantly improve trust, while outcome feedback has a more reliable and positive impact on it. Our interpretation draws on two streams of prior research. Firstly, Hidalgo et al. [45] suggested that while humans judge each other based on intentions, they assess machines by their outcomes. Though interpretability and intention are not identical—interpretability simply explains what factors led an AI system to reach its predictions—Hidalgo et al.'s finding aligns with our observation. Specifically, trust in AI systems (i.e., machines) seems to hinge more on feedback regarding the accuracy of AI outcomes than on information about the underlying rationale for those predictions.

Secondly, our findings correspond with Human-centered Explainable AI (HCXAI) research. While some studies advocate interpretability as a means to increase user trust and performance in AI systems [73], others have pointed out its limitations. Jacobs et al. [48] demonstrated that interpretability doesn't resolve issues such as biased AI recommendations and overreliance on flawed ML algorithms. Krishna et al. [55] further showed that AI-generated explanations often conflict with human knowledge, and even state-of-the-art interpretability methods frequently disagree among themselves. These limitations imply that users might find it challenging to learn from or utilize interpretable AI systems effectively.

However, some scholars argue that the limitations are not inherent to interpretability but arise from current techno-centric perspectives [26, 55, 90]. They propose that adopting a sociotechnical perspective or pursuing human-centered approaches [26, 90] could make interpretability a valuable tool for enhancing trust in AI systems. For example, Park et al. [78] contend that well-designed explanations can boost trust within specific contexts like human resource management, given that various organizational and social factors are considered. Further, studies like that of Chen et al. [18] indicate that considering human cognitive mechanisms in the design of interpretability can also be beneficial.

Future research should concentrate not only on creating more informative explanations but also on devising strategies that ensure these explanations cultivate appropriate trust, considering both behavioral trust and trust calibration perspectives, thereby improve performance. Additionally, identifying algorithmic, social, or human elements that can more directly influence user trust could serve to compensate for the limitations in current interpretability frameworks. Promising areas for future research include designs aimed at improving human-AI collaboration [32, 53], enhancing the controllability of AI systems [24], refining interpretability presentations and interaction methods based on human cognition and contextual needs [18, 20, 39, 41, 43, 59, 71, 71, 94, 96, 104, 106], and bolstering both procedural and social transparency [25, 77, 78].

## 6.4 Trust-Performance Paradox in Outcome Feedback

An important finding from our experiment is that increased trust in AI does not always lead to equally significant improvements in human performance. This observation aligns with existing literature, where previous studies have attempted to explain this phenomenon from a broader perspective, examining the distinctions among trusting beliefs, trusting intentions, and trust-related behaviors [40]. Closely related to our findings, several studies have addressed the trust-performance paradox using more specific concepts: overtrust and undertrust. For example, Jacobs et al. [48] have pointed out the risk of overreliance hampering the performance of AI recommendations. Similarly, some studies have explored this trust-performance paradox through trust calibration, which classifies humans' adoption of AI into overtrust, appropriate trust, and undertrust [75, 89]. We have extended this line of work by disentangling user trust from the model accuracy within the WoA framework. This approach allows us to clarify when and how humans overtrust (i.e., overshoot) or undertrust (i.e., contradict) AI decisions, particularly in relation to outcome feedback and time-dependent behavioral trends.

Additional research regarding how to prevent users from overtrusting and undertrusting AI would be a key future research topic. For example, Fügener et al. [32] have investigated a delegation design that increases benefit to human-AI collaboration compared to either humans or ML algorithms individually. Additionally, any research, even outside the context of interpretability or outcome feedback, that can shed light on the relationship between trust in AI and human performance in a prediction task would be highly significant. Greater clarity regarding when and how trust improvements translate into performance improvements will support not only greater adoption of AI systems but also greater impact from these systems.

## 6.5 Differences between Experts and Lay Users

Any discussion of interpretability should differentiate between experts and lay users, as interpretability is inherently human-centric. These groups vary in their AI literacy and decision-making expertise, which in turn affects their interaction with AI systems [26, 55, 90]. For example, a study focused on data scientists and machine learning practitioners found that these experts tended to overtrust and misuse interpretability tools [52]. Similarly, another study showed that interpretability alone could not improve decision-making accuracy among clinicians, failing to mitigate overreliance on flawed AI suggestions [48]. In contrast, our research, which focuses on lay users, found no significant increase in trust due to interpretability, even though participants reported a strong understanding of both the AI recommendations and the associated explanations (see SI section "Self-Report Measures"). As Wang and Yin [96] suggested, a lack of expertise or AI literacy may be responsible for this finding. Future research that focuses on the roles of AI literacy and expertise could yield valuable insights.

# 7 LIMITATION

Our experiment focuses on a specific context—speed dating predictions—where the decision subjects (i.e., speed dating couples) are distinct from the decision-makers (i.e., participants). This setup parallels many real-world applications of expert AI systems, such as loan officers using AI for loan approvals, doctors using AI for diagnoses, and judges employing AI for sentencing. However, another noteworthy context exists where the subjects of the decisions also have agency in deciding whether to use AI. Examples include individuals deciding AI-generated recommendations tailored specifically for them. The impact of interpretability and outcome feedback on trust may vary between these two contexts. Given this potential variation, future research should evaluate the significance of interpretability and outcome feedback in settings where the subjects of AI predictions also possess decision-making power. Exploring this angle could illuminate how context-specific factors influence the degree of trust placed in AI systems.

Additionally, while our experiment evaluates the presentation variations in UI design, local interpretability (i.e., with or without range condition), and outcome feedback, it explores only a limited range of forms. Specifically, the interpretations in our experiment were presented as lists of factors deemed important by the AI system in making its decision, along with the magnitude of their importance. For local interpretability, we also included information on whether the factor positively or negatively affected the AI system's prediction of a couple's likelihood of a second date. However, there are alternative ways to present interpretability. For example, one could present only the most crucial factors, focus on explanations that are unusual for the decision, or highlight 'what would need to change in the input for the ML prediction/decision to change,' known as 'contrastive' or 'counterfactual' explanations [14]. These and other methods, described by Carvalho et al. [14] and based on research by Breiman [9], Kahneman and Tversky [50], and Lipton [60], warrant further exploration. This area of research regarding how to present interpretations so that they are most beneficial deserves additional attention [18–20, 39, 41, 43, 59, 71, 71, 104]. We believe that the literature needs to focus as much on how to design AI interfaces and present interpretations as it has on techniques for generating interpretations.

# 8 CONCLUSION

Although AI systems excel in various domains, their adoption often faces resistance due to a lack of human trust. Researchers in HCI and social sciences have sought to understand the factors that influence this trust, while computer scientists have grappled with the lack of interpretability in high-performance AI techniques. Despite the prevailing belief that a lack of interpretability may hinder AI adoption, there is insufficient empirical evidence to support this claim. To address this gap, we designed an interactive experiment to examine how interpretability and outcome feedback influence human trust in AI and performance in AI-assisted tasks. Contrary to the prevailing focus on interpretability as a key factor, our findings suggest that outcome feedback may be more effective at fostering trust. Furthermore, our experiment indicates that improving human performance through AI is not solely a matter of increasing trust; higher levels of trust do not necessarily translate into improved human performance.

The literature has delineated two primary categories of factors that influence trust in AI: performance-based factors like model accuracy and outcome feedback, and model-based factors such as interpretability and transparency. Our study is unique in that it directly compares these two categories, focusing specifically on their impact on human trust in AI and, consequently, on user performance. Future research could potentially conduct a comprehensive comparison of all hypothesized factors affecting trust, examining their interplay across various tasks and scenarios. While ambitious, such a study could provide invaluable insights into enhancing trust in AI systems.

# A SUPPLEMENTARY INFORMATION

## A.1 Examples of Phase 1 and Phase 2 Task Instances for Both Experiments 1 and 2

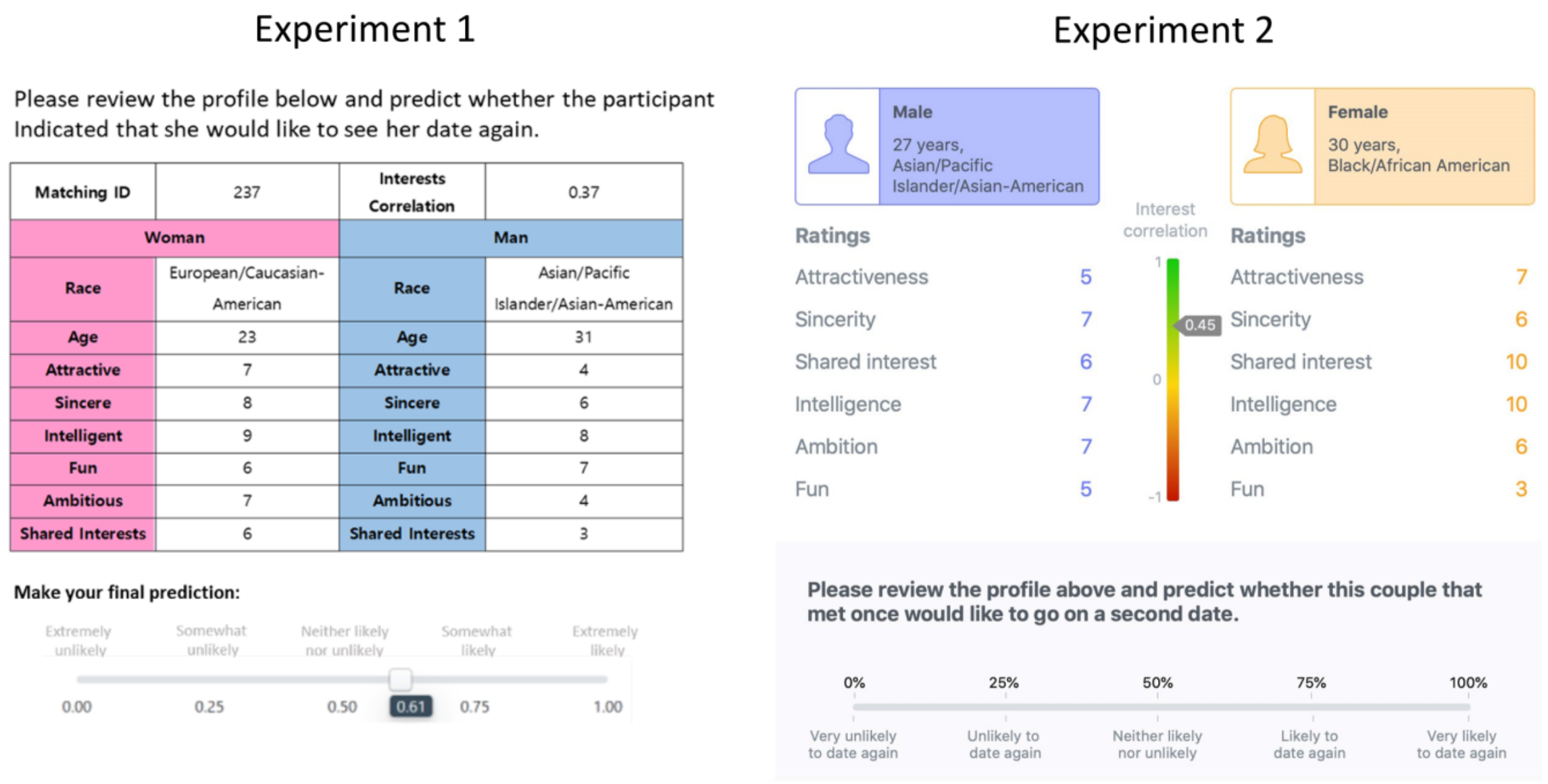

**Figure S1: Phase 1 task instance. Examples of a task instance in phase 1 for experiments 1 and 2.**

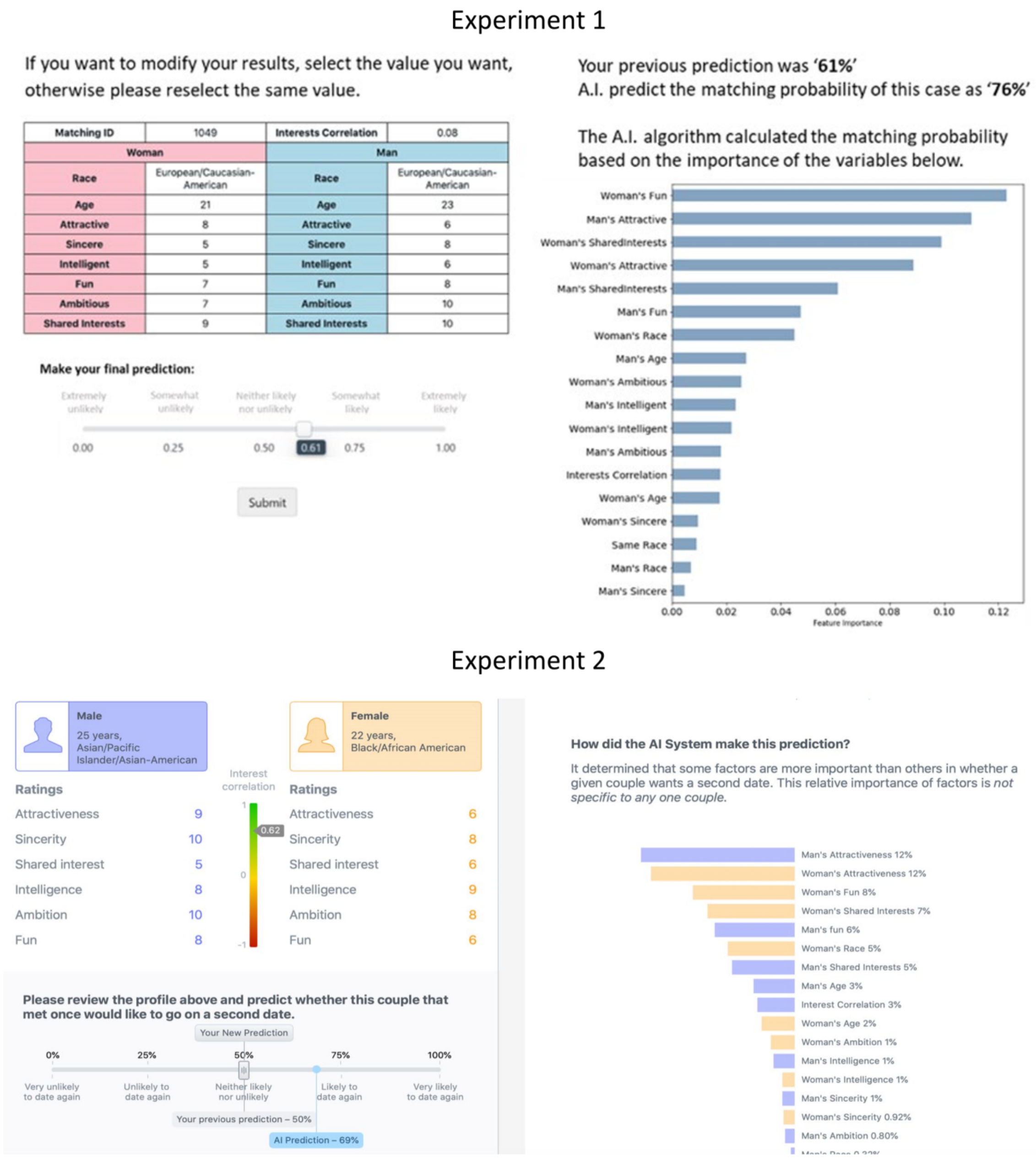

**Figure S2: Phase 2 task instance. Examples of a task instance in phase 2 (shown with global interpretability) for experiments 1 and 2.**

## A.2 Examples of Global Interpretability, Local Interpretability, and Outcome Feedback for Both Experiments 1 and 2

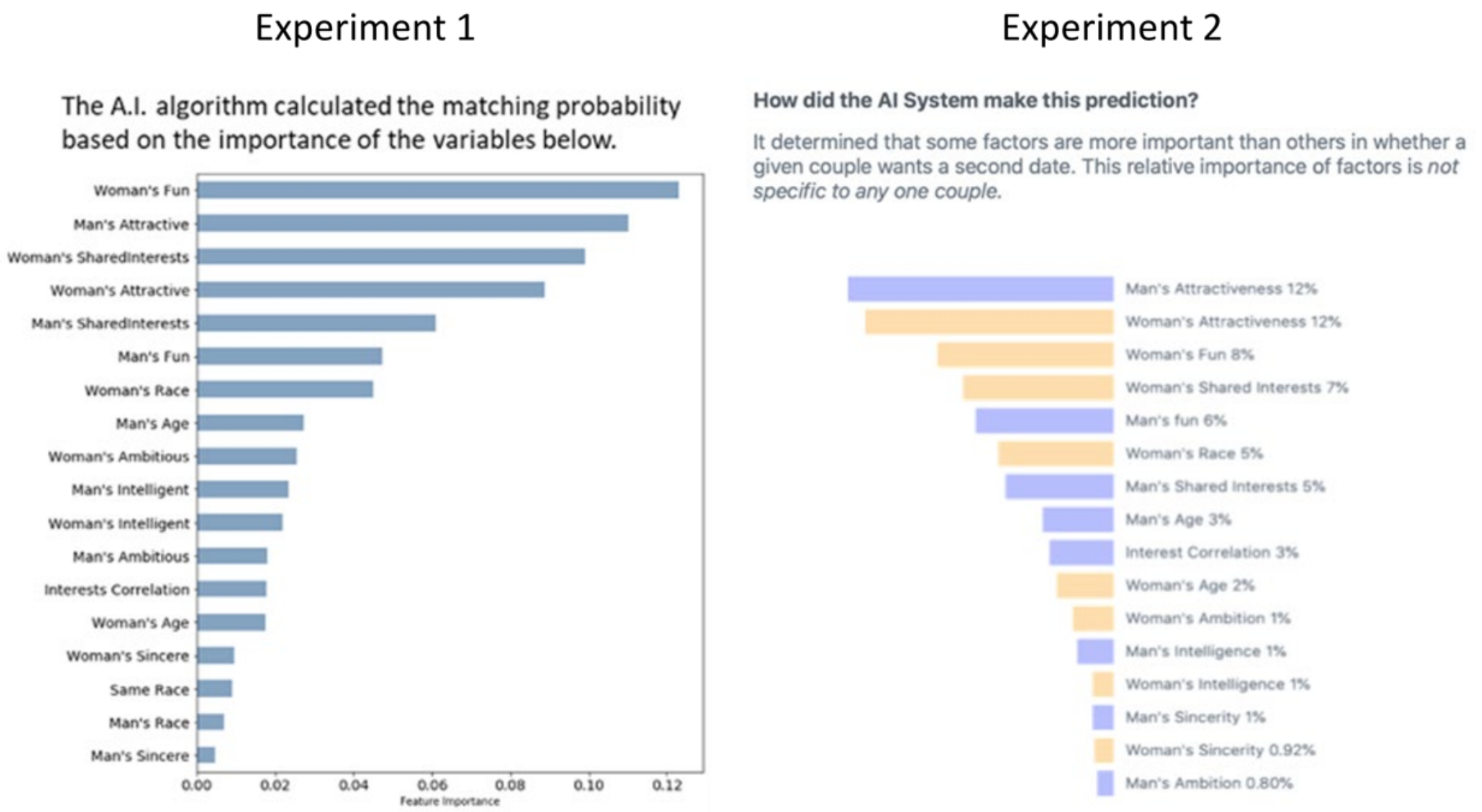

**Figure S3: Global Interpretability. Examples of global interpretability for experiments 1 and 2.**

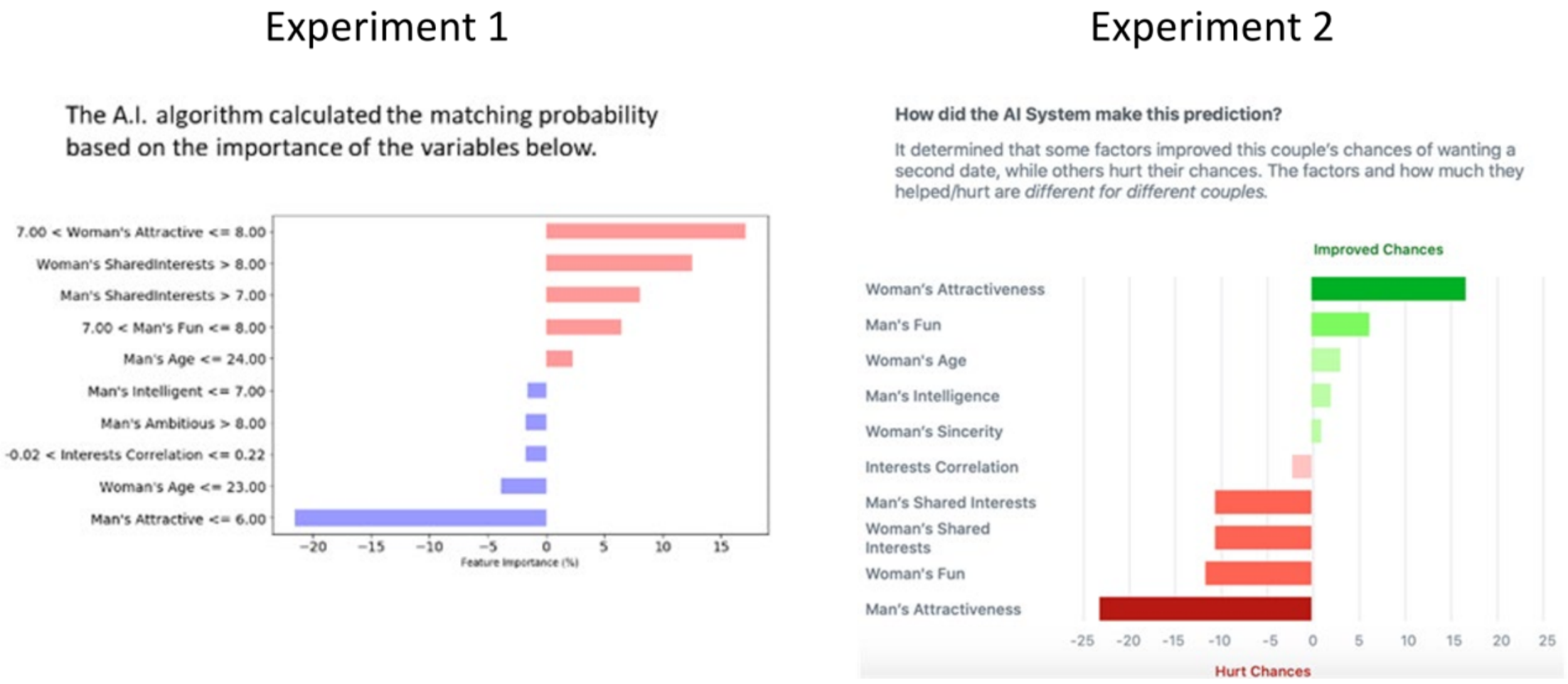

**Figure S4: Local Interpretability. Examples of Local interpretability for experiments 1 and 2.**

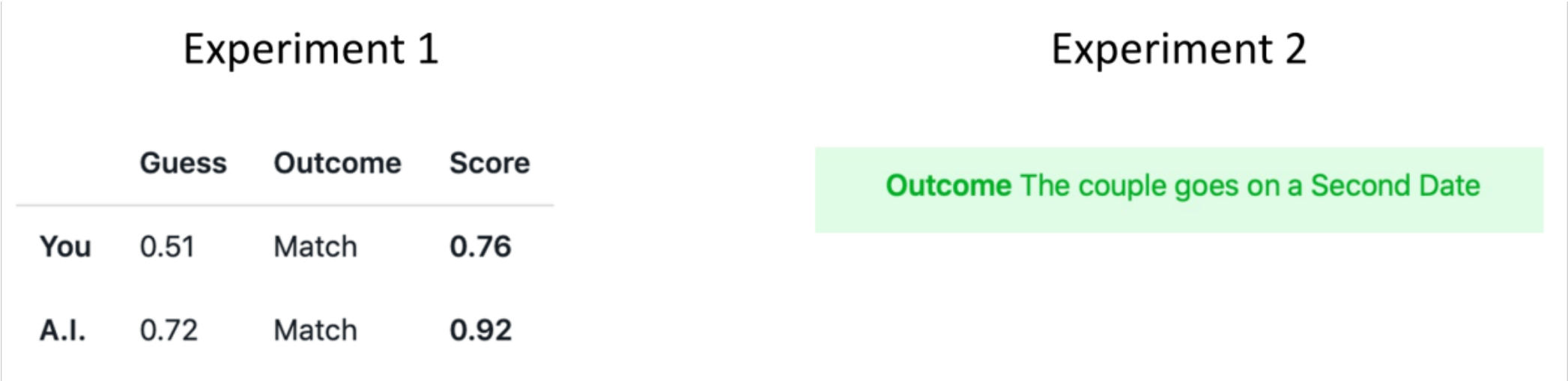

**Figure S5: Outcome Feedback. Examples of outcome feedback, shown for experiments 1 and 2 for the "match" outcome where the couple did go on a second date.**

## A.3 Robustness Checks

Robustness checks for the results presented in the main text are described below, under the following sections "Behavioral Trust," "Performance," and "Time-Dependent Trends."

***Behavioral Trust.*** As described previously, our primary behavioral trust measure (WoA) involved dropping WoA scores when |AI prediction - initial prediction| < 0.15. In addition to the threshold of 0.15, three other thresholds (0.05, 0.1, and 0.2) were used as robustness checks. The results of these robustness checks are displayed in Figure S6 and Table 2 below. As shown in Figure S6 and Table 2, the findings from these robustness checks are consistent with our main findings that outcome feedback led to the greatest and most reliable increase in behavioral trust, while interpretability did not lead to a robust increase in trust. There were also no differences between global and local interpretability, and no interaction between outcome feedback and interpretability, in terms of their impacts on trust.

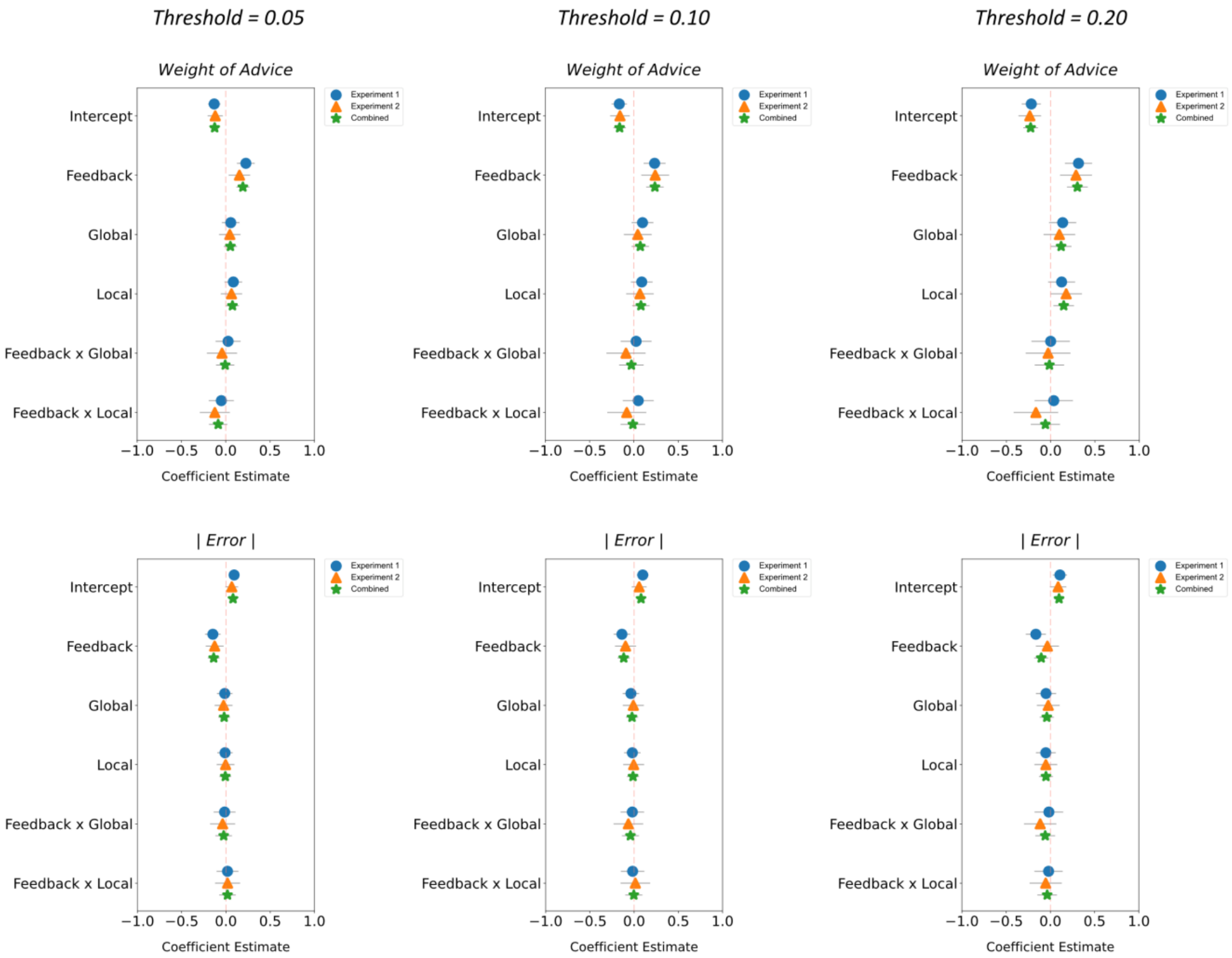

**Figure S6: Robustness checks regarding the impact of outcome feedback and interpretability on behavioral trust. In order to assess the robustness of our primary result, Weight of Advice was calculated using three additional thresholds (0.05, 0.10, and 0.20). The results of these robustness checks are consistent with our main findings.**

**Table S1: Statistics for the robustness checks regarding the impact of outcome feedback and interpretability on behavioral trust. In order to assess the robustness of our primary result, Weight of Advice was calculated using three additional thresholds (0.05, 0.10, and 0.20). For each threshold, the p-values and confidence intervals for each factor are listed. The results of these robustness checks are consistent with our main findings.**

| Y | Exp | Factors | *Threshold = 0.05* | | | *Threshold = 0.10* | | | *Threshold = 0.20* | | |
|---|---|---|---|---|---|---|---|---|---|---|---|
| | | | P-value | CI Lo | CI Hi | P-value | CI Lo | CI Hi | P-value | CI Lo | CI Hi |
| WoA | 1 | Feedback | 0.000 | 0.127 | 0.327 | 0.000 | 0.110 | 0.357 | 0.000 | 0.161 | 0.468 |
| | | Global | 0.268 | -0.043 | 0.156 | 0.122 | -0.026 | 0.220 | 0.084 | -0.018 | 0.289 |
| | | Local | 0.092 | -0.014 | 0.185 | 0.158 | -0.034 | 0.211 | 0.110 | -0.028 | 0.276 |
| | | Feedback×Global | 0.711 | -0.114 | 0.168 | 0.780 | -0.149 | 0.199 | 1.000 | -0.217 | 0.217 |
| | | Feedback×Local | 0.489 | -0.191 | 0.091 | 0.581 | -0.125 | 0.224 | 0.749 | -0.181 | 0.251 |
| | 2 | Feedback | 0.012 | 0.034 | 0.276 | 0.003 | 0.083 | 0.397 | 0.002 | 0.107 | 0.466 |
| | | Global | 0.454 | -0.074 | 0.166 | 0.586 | -0.113 | 0.200 | 0.280 | -0.080 | 0.277 |
| | | Local | 0.292 | -0.055 | 0.184 | 0.392 | -0.088 | 0.223 | 0.053 | -0.002 | 0.353 |
| | | Feedback×Global | 0.627 | -0.212 | 0.128 | 0.429 | -0.310 | 0.132 | 0.823 | -0.281 | 0.224 |
| | | Feedback×Local | 0.154 | -0.292 | 0.046 | 0.473 | -0.301 | 0.139 | 0.195 | -0.417 | 0.085 |
| | 1+2 | Feedback | 0.000 | 0.120 | 0.267 | 0.000 | 0.138 | 0.335 | 0.000 | 0.185 | 0.419 |
| | | Global | 0.158 | -0.020 | 0.126 | 0.149 | -0.026 | 0.170 | 0.048 | 0.001 | 0.235 |
| | | Local | 0.041 | 0.003 | 0.149 | 0.114 | -0.019 | 0.176 | 0.013 | 0.032 | 0.264 |
| | | Feedback×Global | 0.904 | -0.110 | 0.097 | 0.687 | -0.167 | 0.110 | 0.870 | -0.179 | 0.152 |
| | | Feedback×Local | 0.106 | -0.188 | 0.018 | 0.866 | -0.151 | 0.127 | 0.480 | -0.224 | 0.105 |
| Abs Error | 1 | Feedback | 0.001 | -0.236 | -0.060 | 0.004 | -0.232 | -0.043 | 0.005 | -0.278 | -0.050 |
| | | Global | 0.757 | -0.101 | 0.073 | 0.447 | -0.130 | 0.057 | 0.398 | -0.163 | 0.065 |
| | | Local | 0.815 | -0.097 | 0.076 | 0.673 | -0.113 | 0.073 | 0.365 | -0.164 | 0.060 |
| | | Feedback×Global | 0.812 | -0.138 | 0.108 | 0.763 | -0.153 | 0.112 | 0.837 | -0.178 | 0.144 |
| | | Feedback×Local | 0.777 | -0.106 | 0.141 | 0.800 | -0.150 | 0.115 | 0.803 | -0.180 | 0.139 |
| | 2 | Feedback | 0.013 | -0.228 | -0.026 | 0.110 | -0.216 | 0.022 | 0.613 | -0.164 | 0.097 |
| | | Global | 0.602 | -0.128 | 0.074 | 0.885 | -0.128 | 0.110 | 0.714 | -0.154 | 0.105 |
| | | Local | 0.939 | -0.104 | 0.097 | 0.932 | -0.123 | 0.113 | 0.434 | -0.180 | 0.077 |
| | | Feedback×Global | 0.605 | -0.179 | 0.104 | 0.446 | -0.233 | 0.102 | 0.216 | -0.298 | 0.067 |
| | | Feedback×Local | 0.796 | -0.123 | 0.160 | 0.868 | -0.153 | 0.181 | 0.569 | -0.235 | 0.129 |
| | 1+2 | Feedback | 0.000 | -0.205 | -0.072 | 0.001 | -0.186 | -0.049 | 0.011 | -0.183 | -0.024 |
| | | Global | 0.552 | -0.086 | 0.046 | 0.493 | -0.092 | 0.044 | 0.315 | -0.120 | 0.039 |
| | | Local | 0.833 | -0.073 | 0.059 | 0.714 | -0.080 | 0.055 | 0.213 | -0.128 | 0.028 |
| | | Feedback×Global | 0.592 | -0.119 | 0.068 | 0.399 | -0.138 | 0.055 | 0.307 | -0.170 | 0.054 |
| | | Feedback×Local | 0.709 | -0.075 | 0.111 | 0.949 | -0.099 | 0.093 | 0.522 | -0.147 | 0.075 |

***Performance.*** In addition to the primary performance measure of absolute error, three additional measures were used as robustness checks, including squared error, square root error, and the area under the ROC curve. In our experiments, these three measures were calculated in the following way:

$$\text{Squared Error} = (actual\ value\ -\ revised\ prediction)^2$$

$$\text{Square Root Error} = \sqrt{(actual\ value\ -\ revised\ prediction)}$$

ROC AUC measures the two-dimensional area underneath the ROC curve. The ROC curve is a graph representing the performance of a classification model at all thresholds. This curve plots two parameters: True Positive Rate (TPR) on the y-axis and False Positive Rate (FPR) on the x-axis. TRP can be computed as {(True Positive) / (True Positive + False Negative)}. FPR is computed as {(False Positive) / (False Positive + True Negative)}. True and false indicate the real value in the classification task. For example, in our experiment, true and false mean 'match' and 'no match', respectively. Positive and negative are related to the participant's prediction (when the participant predicts "match" it is positive, and when the participant predicts "no match" it is negative). In addition to the primary performance measure of absolute error, three additional measures were used as robustness checks, including squared error, square root error, and the area under the ROC curve.

Results for these robustness checks are displayed in Figure S7 and Table 3 below. As shown in Figure S7 and Table 3, the results for squared error and square root error are directionally consistent with our main findings that feedback increases performance (in experiment 1 and in the combined dataset), though neither interpretability nor the interaction between feedback and interpretability impact performance. While the results for squared error and square root error are directionally consistent with these findings, the size of the impact is smaller for squared error and larger for square root error due to the way these measures are calculated. Specifically, squared error tends to amplify the effect of large errors, while square root error minimizes the effect of large errors. Results from our experiment suggest that outcome feedback increased participants' tendency to make large errors (by leading participants to make more extreme predictions, sometimes "contradicting" and sometimes "overshooting" the AI advice), which has resulted in a smaller increase in performance seen in the squared error measurement and a larger increase in performance seen in the square root error measurement (as compared to the primary measure of absolute error).

With regards to the ROC AUC measure, it is important to note that ROC AUC does not measure performance by measuring error, meaning that the direction of the ROC AUC measure is reverse to the error measures (higher ROC AUC indicates improved performance, whereas higher error indicates decreased performance). Furthermore, a critical difference between ROC AUC and measures of error is that ROC AUC is calculated at the level of participants, as opposed to at the level of individual task instances. As a result, ROC AUC is an unstable measure of performance in this experiment, as shown in Figure S7 and Table 10. This is due to the relatively small number of task instances in our experiment. Because there were only ten task instances for each participant (the first two of the twelve total task instances were for practice purposes), measuring performance at the participant level was not particularly stable or meaningful. Getting an accurate measure of performance at the participant level would have required a significantly greater number of task instances per participant. As such, despite pre-registering ROC AUC as our performance measure, we instead used absolute error as the primary performance measure (with squared error and square root error as the main robustness checks).

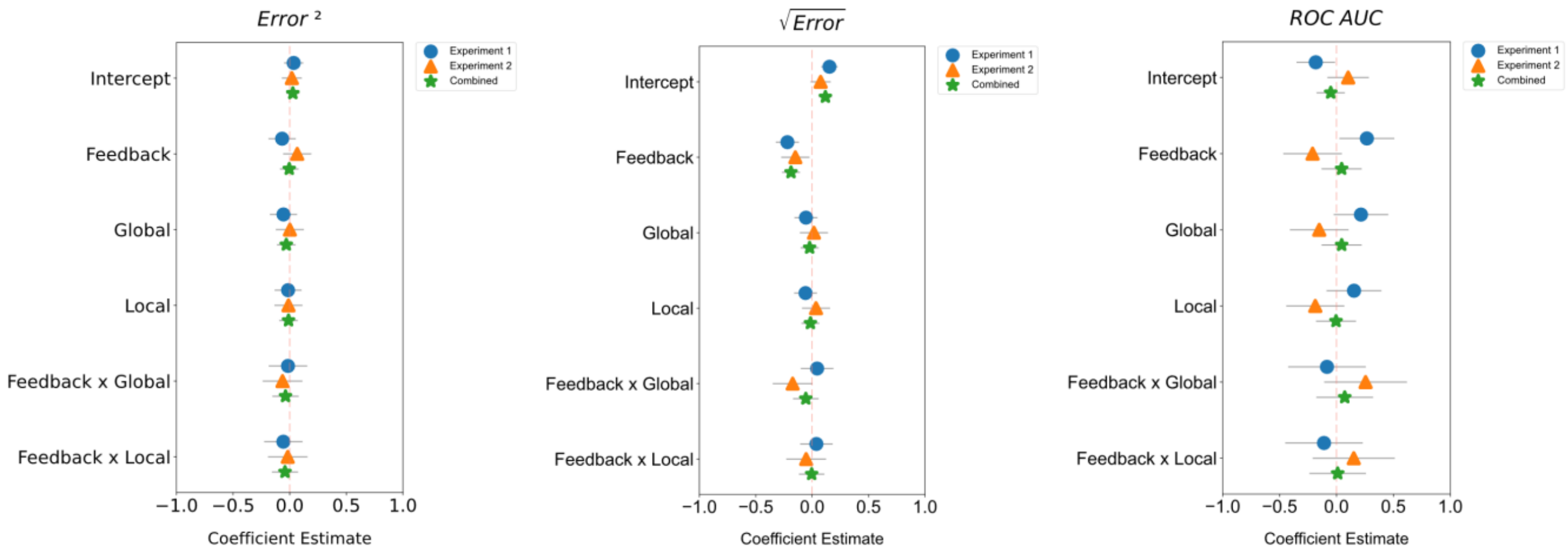

**Figure S7: Robustness checks regarding the impact of outcome feedback and interpretability on performance. Squared error, square root error, and ROC AUC were used to assess the robustness of our primary result (calculated using absolute error). The results for squared error and square root error are directionally consistent with our main findings, though the size of the impact is smaller for squared error and larger for square root error due to the way these measures are calculated. The results for ROC AUC were unstable. ROC AUC measures performance at the participant level instead of at the task instance level, and this turned out to be an unstable way to measure performance in this experiment given the relatively small number of task instances in our experiment.**

**Table S2: Statistics for the robustness checks regarding the impact of outcome feedback and interpretability on performance. Squared error, square root error, and ROC AUC were used to assess the robustness of our primary result (calculated using absolute error). For each measure the p-values and confidence intervals for each factor are listed. The results for squared error and square root error are directionally consistent with our main findings, while the results for ROC AUC are unstable as ROC AUC measures performance at the participant level, not at the task instance level.**

| Exp | Factors | *Squared Error* | | | *Square Root Error* | | | *ROC AUC* | | |
|---|---|---|---|---|---|---|---|---|---|---|
| | | P-value | CI Lo | CI Hi | P-value | CI Lo | CI Hi | P-value | CI Lo | CI Hi |
| 1 | Feedback | 0.277 | -0.187 | 0.054 | 0.000 | -0.322 | -0.117 | 0.030 | 0.026 | 0.509 |
| | Global | 0.371 | -0.175 | 0.065 | 0.292 | -0.157 | 0.047 | 0.079 | -0.025 | 0.455 |
| | Local | 0.813 | -0.134 | 0.105 | 0.254 | -0.160 | 0.042 | 0.209 | -0.087 | 0.395 |
| | Feedback×Global | 0.862 | -0.185 | 0.155 | 0.554 | -0.101 | 0.189 | 0.631 | -0.424 | 0.257 |
| | Feedback×Local | 0.517 | -0.226 | 0.113 | 0.611 | -0.107 | 0.182 | 0.528 | -0.451 | 0.231 |
| 2 | Feedback | 0.294 | -0.058 | 0.191 | 0.020 | -0.275 | -0.024 | 0.110 | -0.468 | 0.048 |
| | Global | 0.982 | -0.123 | 0.125 | 0.807 | -0.109 | 0.140 | 0.249 | -0.409 | 0.106 |
| | Local | 0.863 | -0.134 | 0.113 | 0.593 | -0.091 | 0.158 | 0.152 | -0.441 | 0.069 |
| | Feedback×Global | 0.478 | -0.239 | 0.112 | 0.055 | -0.350 | 0.003 | 0.167 | -0.108 | 0.619 |
| | Feedback×Local | 0.846 | -0.192 | 0.158 | 0.553 | -0.230 | 0.123 | 0.409 | -0.209 | 0.513 |
| 1+2 | Feedback | 0.922 | -0.087 | 0.078 | 0.000 | -0.268 | -0.108 | 0.614 | -0.131 | 0.222 |
| | Global | 0.472 | -0.112 | 0.052 | 0.577 | -0.103 | 0.057 | 0.611 | -0.130 | 0.222 |
| | Local | 0.801 | -0.092 | 0.071 | 0.696 | -0.095 | 0.064 | 0.964 | -0.180 | 0.172 |
| | Feedback×Global | 0.533 | -0.153 | 0.079 | 0.327 | -0.170 | 0.057 | 0.568 | -0.177 | 0.322 |
| | Feedback×Local | 0.481 | -0.158 | 0.074 | 0.937 | -0.117 | 0.108 | 0.933 | -0.238 | 0.259 |

***Time-Dependent Trend.*** In addition to the primary measure of absolute error that was used in the analysis of the time-dependent trend regarding performance, two additional measures (squared error and square root error) were also used as robustness checks. The results for these robustness checks were directionally consistent with our main finding that providing outcome feedback generally increases performance over time, but can reduce it after cases in which AI "harmed" the participant.

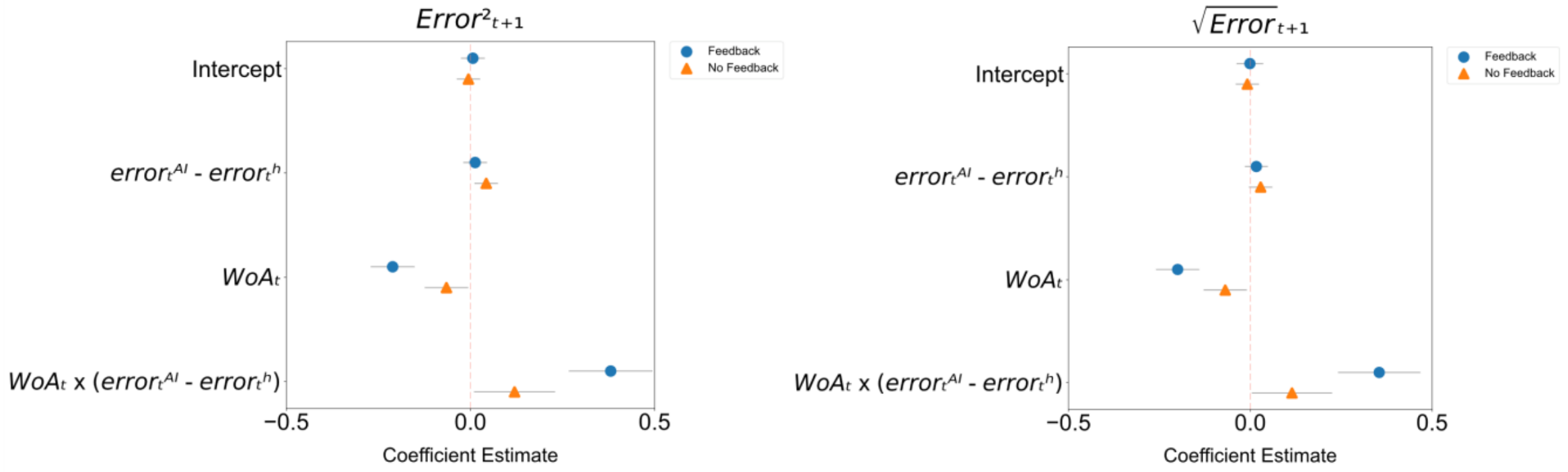

**Figure S8: The robustness checks regarding the time-dependent trends on performance. Squared error and square root error were used to assess the robustness of our time-dependent trend of performance. The results for squared error and square root error are directionally consistent with our main finding, as providing feedback generally increases performance over time, but can reduce it when AI "harmed" the participant.**

## A.4 Self-report Measure

After completing the experiment, all participants in all conditions were asked about the ease with which they could understand the information provided by the AI system. Specifically, participants were asked, "Having experienced the AI system, was it easy to understand?" and they responded on a 5-point likert scale (where 1 = extremely difficult, 2 = somewhat difficult, 3 = neither easy nor difficult, 4 = somewhat easy, and 5 = extremely easy). Results from this question for each of the two experiments are shown in Figure S8. The average is above 4 (between "somewhat easy" and "extremely easy") for all conditions in both experiments, except for the outcome feedback + local interpretability condition in the first experiment, where the average is slightly below 4 (between "neither easy nor difficult" and "somewhat easy"). As such, participants did not indicate that they had difficulty understanding the AI system regardless of condition.

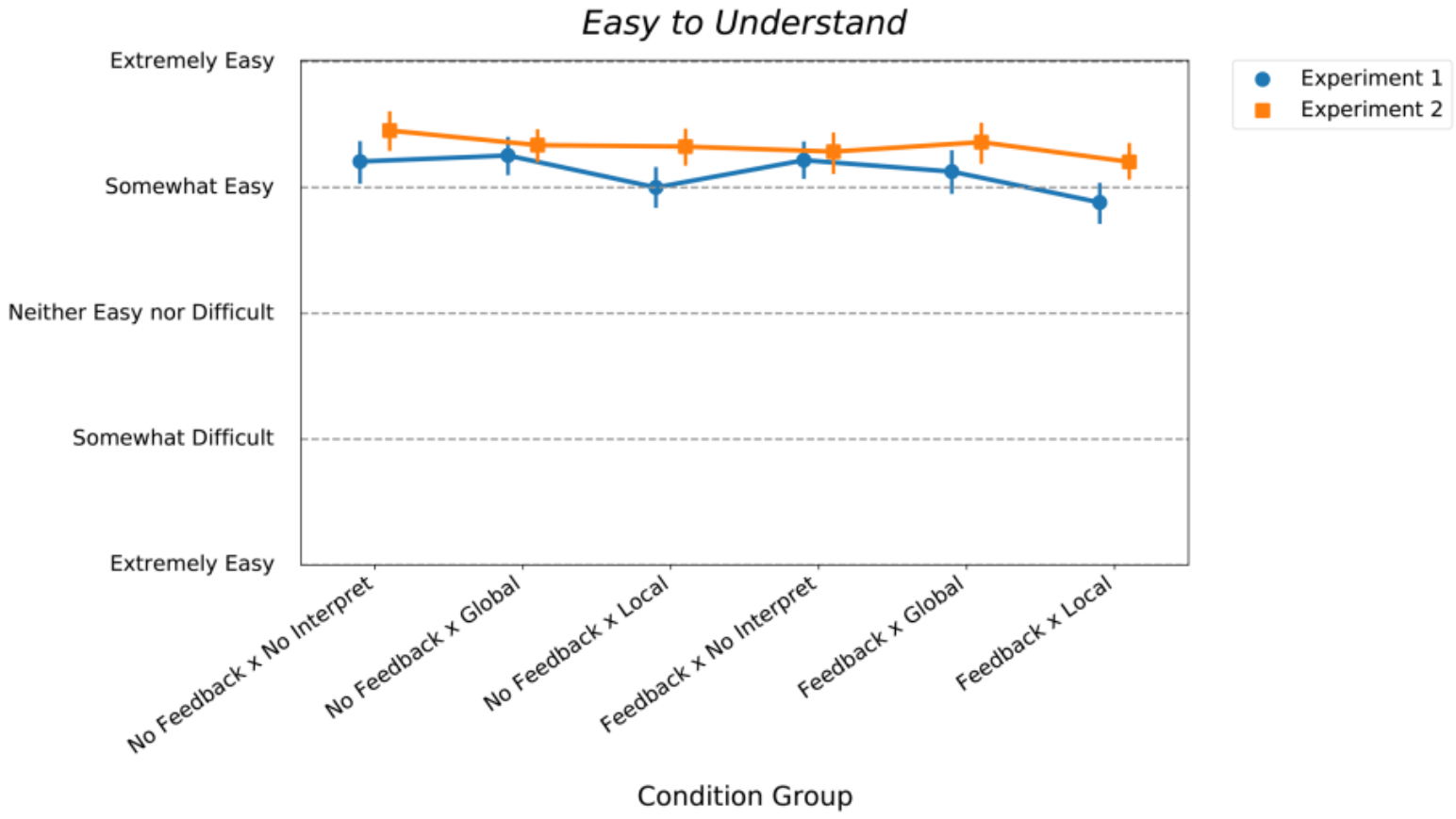

**Figure S9: The self-reported ease with which participants could understand the information provided by the A.I. system. Participants were asked, "Having experienced the A.I. system, was it easy to understand?" and they responded on a 5-point likert scale (where 1 = extremely difficult, 2 = somewhat difficult, 3 = neither easy nor difficult, 4 = somewhat easy, and 5 = extremely easy). The average scores are shown for each condition in each of the two experiments, with experiment 1 depicted in blue dots and experiment 2 depicted in orange squares. The average scores were between "somewhat easy" and "extremely easy," for all conditions across both experiments except the feedback + local interpretability condition in experiment 1, for which the average score was between "neither easy nor difficult" and "somewhat easy."**

## A.5 Pre-registered Hypotheses

**CONFIDENTIAL - FOR PEER-REVIEW ONLY**

**Impact of Model Interpretability and Performance Feedback (#37908)**

**Created:** 03/24/2020 03:58 PM (PT)

This is an anonymized copy (without author names) of the pre-registration. It was created by the author(s) to use during peer-review.
A non-anonymized version (containing author names) should be made available by the authors when the work it supports is made public.

**1) Have any data been collected for this study already?**
No, no data have been collected for this study yet.

**2) What's the main question being asked or hypothesis being tested in this study?**
We have two main hypotheses:

The first hypothesis is that "model interpretability" and "performance feedback" increase the user's trust in the model prediction. We expect an interaction between model interpretability and performance feedback where providing feedback will be most effective when no model interpretation is provided

The second hypothesis is that "model interpretability" and "performance feedback" will increase the prediction accuracy of the human judge. We expect the model performance, on average, will be superior to the human judge, therefore, the increased trust in the model will lead to higher human judge performance.

**3) Describe the key dependent variable(s) specifying how they will be measured.**
For the first hypothesis, the dependent variable is "trust," which we operationalize the Weight of Advice (WOA) measure frequently used in the literature on advice-taking. WOA quantifies the degree to which people update their beliefs (e.g., predictions made before seeing the model predictions) toward advice they are given (the model prediction). In the context of our experiments, it is defined as WOA = | u_2 - u_1| / | m - u_1| , where m is the model prediction, u_1 is the participant's initial prediction before seeing m, and u_2 is the participant's final prediction after seeing m. All values of | m - u_1| < 0.05 will be removed (i.e., the initial prediction and model prediction are very similar, so the weight of advice can't be computed).

For the second hypothesis, we operationalize the area under the ROC curve (AUC) of the participant's final prediction after seeing the model prediction.

We will use the standardized score (z-score) of each outcome (i.e., trust and performance) as our dependent variable.

**4) How many and which conditions will participants be assigned to?**
Participants will be assigned to one of six conditions (between-subjects design) with three interpretability conditions (no interpretation, global and local interpretation) that are further partitioned into two feedback conditions (no feedback regarding model performance and with feedback).

**5) Specify exactly which analyses you will conduct to examine the main question/ hypothesis.**
For our primary analyses, we will use a generalized mixed-effects linear model predicting each standardized outcome (i.e., "trust" for hypothesis 1; and "performance" for hypothesis 2) with a dummy variable indicating the condition. The regression will include a subject-level random effect to account for the nested structure of the data. All statistics will be two-tailed.

**6) Describe exactly how outliers will be defined and handled, and your precise rule(s) for excluding observations.**
We will NOT analyze the data that we collect in the practice task (during the instructions). We will also exclude data in a case by a particular subject if no initial prediction is submitted for that case. For our first hypothesis, we will exclude all observations with values of | m - u_1| < 0.05 (i.e., the initial prediction and model prediction are very similar, so the weight of advice can't be computed).

**7) How many observations will be collected or what will determine sample size? No need to justify decision, but be precise about exactly how the number will be determined.**
We will recruit 800 participants in total. All participants will be recruited from Amazon Mechanical Turk.

**8) Anything else you would like to pre-register? (e.g., secondary analyses, variables collected for exploratory purposes, unusual analyses planned?)**
For robustness checks, we use additional measures for each hypothesis. For the first hypothesis, the absolute change in the participant's belief will be used. That is, we will calculate the unsigned difference between the two predictions from participants (before and after seeing the model prediction). For the second hypothesis, we operationalize performance as the relative improvement (before and after seeing the model prediction) in AUC.

Also, we include a battery of questions for exploratory purposes and robustness checks of our operationalizations, including self-reported trust, self-reported willingness to incorporate model predictions into their decision-making, social perceptiveness scores, and demographics information. Exploratory such analyses could include investigating heterogeneity in treatment effects as well as sensitivity to different operationalizations of our dependent variables.

Version of AsPredicted Questions: 2.00

**Available at** https://aspredicted.org/JEY_UJP